\documentclass{article}
\usepackage{log_2026}                  

\usepackage{tabularx}
\usepackage{array}
\usepackage{booktabs}
\usepackage{multirow}
\usepackage{amsfonts}
\usepackage{amsmath}
\usepackage{amssymb}
\usepackage{graphicx}
\usepackage{subcaption}
\usepackage{xcolor}
\usepackage{tikz}
\usepackage{wrapfig}
\usetikzlibrary{arrows.meta, positioning, calc, shapes.geometric}
\usepackage[numbers,compress,sort]{natbib}

\usepackage{caption}
\newcommand{\R}{\mathbb{R}}
\newcommand{\sre}{\mathrm{SRE}}

\definecolor{TrajPurple}{RGB}{106,61,154}
\definecolor{VarBlue}{RGB}{31,119,180}
\definecolor{ConsTeal}{RGB}{0,128,128}
\definecolor{TrajGreen}{RGB}{44,160,44}
\definecolor{PosterNavy}{RGB}{27,54,93}

\title[Learning SOCP Solver Dynamics with Graph Neural Networks]
{Learning SOCP Solver Dynamics with Graph Neural Networks}

\author[A. Ghalem et al.]{%
Adam Ghalem$^{1,2}$ \quad Lou Salaün$^{2}$ \quad Davide Rinaldi$^{2}$ \quad Fragkiskos D. Malliaros$^{1}$\\
\normalfont
$^{1}$Universit\'e Paris-Saclay, CentraleSup\'elec, Inria \qquad
$^{2}$Nokia Bell Labs\\
\email{adam.ghalem@nokia.com} \quad \email{fragkiskos.malliaros@centralesupelec.fr}\\
\email{\{lou.salaun, davide.rinaldi\}@nokia-bell-labs.com}
}
\begin{document}

\maketitle

\begin{abstract}
Second-order cone programs (SOCPs) are a fundamental class of convex optimization problems widely applied across engineering domains. Classical interior-point methods are robust and accurate, but their iterative nature can be costly in real-time or repeated-solve settings. This has motivated learning-based solvers that approximate optimizer solutions with neural networks. Recent work has increasingly studied graph neural networks (GNNs) for linear, quadratic, and second-order cone optimization, leveraging their ability to model sparse relational structure among variables, constraints, and objective functions.
Most existing GNN-based solvers compress the optimization process into a single forward pass, predicting the optimizer directly without using the intermediate trajectory produced by classical solvers. We introduce SOCPNet, a recursive GNN framework that uses solver trajectories to learn an iterative SOCP solution procedure. A shared GNN model is applied repeatedly, refining the current iterate at each step and enabling inference for a variable number of iterations. Rather than predicting the solution in a single pass, SOCPNet learns from intermediate iterates to progressively refine its estimates using a trajectory-regularized training procedure.
We evaluate SOCPNet on multiple datasets spanning standard SOCP problem classes and show that it consistently reduces the final solution error relative to a SOCP-GNN baseline. Finally, we demonstrate how trained SOCPNet models can be used in a hybrid optimization pipeline by warm-starting a classical interior-point solver, reducing solver iterations without loss of accuracy.
 
\end{abstract}

\section{Introduction}
\label{sec:intro}

Second-order cone programming (SOCP) is a fundamental class of convex optimization
problems underlying a wide range of engineering applications, including wireless
communications and signal processing~\citep{boyd2004convex,lobo1998applications}.
Traditional SOCP solvers rely on interior-point methods
(IPMs)~\citep{nocedal2006numerical,wachter2006implementation}, which are robust and
accurate but iterative and computationally demanding, limiting their use in real-time~\citep{salaun2022gnn}
or repeated-solve settings. This has motivated learning-based
optimization, where graph neural networks (GNNs) exploit the sparsity and
variable-constraint structure of the problem
\citep{bengio2021machine,gasse2019exact,ibarz2022generalist,kipf2017semi,velickovic2018graph},
with applications to SOCP-specific tasks such as power control in telecommunication
networks~\citep{salaun2022gnn,parlier2024learning} and to SOCP subclasses such as linear and quadratically constrained quadratic programs~\citep{qian2024exploring,qian2025towards,wu2024representing}.

\begin{figure}[t]
\centering
\resizebox{\linewidth}{!}{%
\begin{tikzpicture}[
  font=\footnotesize,
  gt/.style={draw=TrajPurple,fill=TrajPurple!10,line width=1pt,rounded corners=5pt,
             minimum width=2.1cm,minimum height=2.0cm,align=center,
             text=TrajPurple!60!black,font=\small\bfseries},
  emb/.style={draw=VarBlue,fill=VarBlue!12,line width=0.8pt,rounded corners=3pt,
              minimum width=0.62cm,minimum height=1.55cm,align=center,font=\tiny,
              text=VarBlue!60!black},
  gnn/.style={draw=ConsTeal,fill=ConsTeal!14,line width=0.9pt,rounded corners=4pt,
              minimum width=1.45cm,minimum height=1.55cm,align=center,
              text=ConsTeal!55!black,font=\small\bfseries},
  rdo/.style={draw=TrajGreen!85!black,fill=TrajGreen!14,line width=0.8pt,rounded corners=3pt,
              minimum width=0.62cm,minimum height=1.55cm,align=center,font=\tiny,
              text=TrajGreen!45!black},
  arr/.style={-{Stealth[length=5pt]},line width=0.8pt,color=PosterNavy},
  lab/.style={font=\scriptsize\bfseries,color=PosterNavy,inner sep=1.5pt},
  loss/.style={draw=PosterNavy,fill=PosterNavy!8,line width=0.8pt,rounded corners=4pt,
               align=center,inner sep=6pt,text=PosterNavy},
  sup/.style={-{Stealth[length=4pt]},dotted,line width=0.9pt,color=PosterNavy}
]
\node[gt] (gt) {Ground-truth\\IPM\\trajectory\\[2pt]$\mathbf{x}^{(0)}\!,\dots,\mathbf{x}^{(T)}$};
\node[emb, right=1.1cm of gt] (e1) {\rotatebox{90}{Embed}};
\node[gnn, right=0.32cm of e1] (g1) {SOCPNet};
\node[rdo, right=0.32cm of g1] (r1) {\rotatebox{90}{Readout}};
\node[emb, right=1.2cm of r1] (e2) {\rotatebox{90}{Embed}};
\node[gnn, right=0.32cm of e2] (g2) {SOCPNet};
\node[rdo, right=0.32cm of g2] (r2) {\rotatebox{90}{Readout}};
\node[font=\large\bfseries, color=PosterNavy, right=0.55cm of r2] (dots) {$\cdots$};
\node[emb, right=0.55cm of dots] (eT) {\rotatebox{90}{Embed}};
\node[gnn, right=0.32cm of eT] (gT) {SOCPNet};
\node[rdo, right=0.32cm of gT] (rT) {\rotatebox{90}{Readout}};

\draw[arr] (gt) -- node[lab,above]{$\mathbf{x}^{(0)}$} (e1);
\draw[arr] (e1) -- (g1); \draw[arr] (g1) -- (r1);
\draw[arr] (r1) -- node[lab,above]{$\hat{\mathbf{x}}^{(1)}$} (e2);
\draw[arr] (e2) -- (g2); \draw[arr] (g2) -- (r2);
\draw[arr] (r2) -- node[lab,above]{$\hat{\mathbf{x}}^{(2)}$} (dots);
\draw[arr] (dots) -- (eT);
\draw[arr] (eT) -- (gT); \draw[arr] (gT) -- (rT);
\draw[arr] (rT) -- ++(1.0,0) node[lab,right]{$\hat{\mathbf{x}}^{(T)}$};

\draw[arr, TrajPurple, dashed, line width=1pt]
  (g1.north) -- ++(0,0.55) -| (g2.north);
\draw[arr, TrajPurple, dashed, line width=1pt]
  (g2.north) -- ++(0,0.55) -| (gT.north);
\node[font=\scriptsize\bfseries, color=TrajPurple!80!black]
  at ($(g1.north)!0.5!(g2.north)+(0,0.85)$) {shared weights};

\node[loss, below=1.05cm of g2] (L)
  {Per-step supervision: trajectory or target};
\draw[sup] (r1.south) -- ++(0,-0.35) -| (L.north);
\draw[sup] (r2.south) -- ++(0,-0.35) -| (L.north);
\draw[sup] (rT.south) -- ++(0,-0.35) -| (L.north);
\draw[sup] (gt.south) |- (L.west);
\end{tikzpicture}
}%
\caption{A single shared SOCPNet is applied recursively.}
\label{fig:trajlearn}
\end{figure}
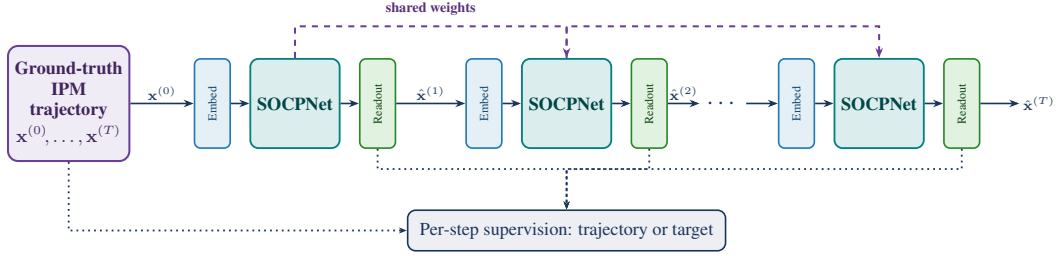

Most existing learning-based solvers adopt an \emph{end-to-end} formulation in which a
GNN predicts the optimal solution $\mathbf{x}^\star$ in a single forward pass. This
compresses the solver dynamics into a fixed-depth computation and discards the
intermediate iterates generated by classical solvers during training-data construction.
In contrast, we train a GNN to recursively refine an SOCP iterate using information from the IPM trajectory
$\{\mathbf{x}^{(0)}, \ldots, \mathbf{x}^{(T)}\}$
(Fig.~\ref{fig:trajlearn}). Our empirical hypothesis is that recasting SOCP solving as recursive solution refinement yields more accurate predictions than direct endpoint regression, by decomposing the solution process into a sequence of learned local updates. Moreover, the learned trajectory is operationally useful: it produces increasingly refined, approximately feasible iterates that can be used
to warm-start an exact solver.

Our objective is to train a GNN-based solver that, given an SOCP instance, returns an
iterate that \textit{(i)} approximates the optimal solution $\mathbf{x}^\star$ and
\textit{(ii)} satisfies the problem constraints to high accuracy.

Unlike standard algorithm unrolling, which typically maps a fixed number of optimization iterations
to a fixed-depth network with distinct or layer-specific parameters \citep{li2025expressivity}, here we introduce SOCPNet which reuses a single
shared update model for all iterations. This shared recursion allows the model to be evaluated for more steps at
inference than were used during training. It also requires the learned update to remain
stable under repeated application. In practice, a naively trained recursive model is not
contractive: when unrolled beyond the training horizon, its iterates may stop improving
and \emph{drift} away from the solution (Section~\ref{sec:recipe}). We therefore introduce
training and architectural components that explicitly promote stable long-horizon
refinement.

Because the learned recursion is computationally efficient but approximate, we further use
SOCPNet as a warm-start mechanism for IPOPT~\citep{wachter2006implementation}, a widely
used interior-point solver for constrained nonlinear programming. After a small number of
learned refinement steps, the SOCPNet iterate is passed to IPOPT as the initial point,
placing the solver closer to the optimum and reducing the number of iterations required
for high-accuracy convergence (Section~\ref{sec:hybrid}).

\textbf{Contributions.} \textit{(i)} We formulate SOCP solving as \emph{recursive}
trajectory learning and introduce SOCPNet, a single shared GNN update applied for a
variable number of steps to approximate an IPM trajectory rather than only its endpoint.
\textit{(ii)} We develop a stability-oriented training procedure for long-horizon
recursive inference, combining residual updates, step-size control, trajectory and
endpoint supervision, KKT-informed features, and feasibility/objective losses.
\textit{(iii)} We evaluate SOCPNet on diverse SOCP datasets, and provide an ablation study
to show that a naively trained recursion drifts without the proposed procedure.
\textit{(iv)} We demonstrate that SOCPNet can warm-start IPOPT, reducing interior-point iterations by $33\%-45\%$ without degrading final solution accuracy.
Code is available at \url{https://github.com/nokia/SOCPNet}.

\section{Problem Formulation and Related Work}
\label{sec:problem}

We consider the standard second-order cone program (SOCP)
\begin{align}
  \text{minimize} \quad & \mathbf{f}^\top \mathbf{x} \label{eq:obj}\\
  \text{subject to} \quad
  & \|\mathbf{A}_k \mathbf{x} + \mathbf{b}_k\|_2
  \leq \mathbf{c}_k^\top \mathbf{x} + d_k, \quad k \in [m], \label{eq:cone}\\
  & \mathbf{F}\mathbf{x} \leq \mathbf{g}, \qquad
  \mathbf{l} \leq \mathbf{x} \leq \mathbf{r}, \label{eq:lin}
\end{align}
over decision variables $\mathbf{x}\in\R^n$~\citep{lobo1998applications}.
The vector $\mathbf{f}\in\R^n$ defines the linear objective. Each conic constraint
$k\in[m]$ is specified by
$\mathbf{A}_k\in\R^{n_k\times n}$,
$\mathbf{b}_k\in\R^{n_k}$,
$\mathbf{c}_k\in\R^n$, and $d_k\in\R$, and constrains the norm
of an affine function of $\mathbf{x}$ by an affine upper bound. The remaining constraints
consist of $p$ linear inequalities, with
$\mathbf{F}\in\R^{p\times n}$, $\mathbf{g}\in\R^p$ and
element-wise lower and upper bounds
$\mathbf{l},\mathbf{r}\in\R^n$.

Interior-point methods (IPMs) solve \eqref{eq:obj}--\eqref{eq:lin} by approximately
satisfying the Karush-Kuhn-Tucker (KKT) conditions: stationarity of the Lagrangian,
primal feasibility, dual feasibility, and complementary slackness between constraints and
their associated multipliers~\citep{nocedal2006numerical}. They do so iteratively, relaxing
the complementarity conditions by a barrier parameter that is decreased to zero, and
produce a sequence of iterates that progressively reduce primal-dual residuals
$\{\mathbf{x}^{(0)},\ldots,\mathbf{x}^{(T)}\}$ and converges to the optimum,
$\mathbf{x}^\star$. We take this sequence as our supervision signal. Details about the
specific solver and termination tolerances used to generate these trajectories are
reported in Appendix~\ref{app:training}.

\noindent\textbf{Related work.}
A growing line of work studies the representational power of GNNs for optimization
problems of increasing generality, including linear
programs~\citep{qian2024exploring}, convex quadratically constrained quadratic
programs~\citep{wu2024representing}, broader classes of quadratic
programs~\citep{qian2025towards}\, and second-order cone
programs~\mbox{\citep{li2025expressivity}}. SOCPs are a natural target for learned optimization
because they include linear programs and many widely used convex quadratic and
quadratically constrained formulations through standard conic reformulations.

GNNs have also been used to learn solution mappings for optimization
problems~\citep{bengio2021machine,gasse2019exact}. Closest to our setting is the
heterogeneous SOCP-GNN of \citet{li2025expressivity}, which predicts the optimal
solution $\mathbf{x}^\star$ in a single forward pass. We retain this graph representation
and its message passing operators, but replace one-shot prediction with a recursive
trajectory-learning formulation that applies the same learned update repeatedly to model
the IPM solution path (Section~\ref{sec:arch}). Our formulation is related to algorithm
unrolling, which maps a fixed number of iterations of an optimization algorithm to a
fixed-depth neural architecture, for example in sparse coding~\citep{gregor2010learning},
ADMM~\citep{yang2016admm}, and primal-dual methods~\citep{chen2018theoretical}. In
contrast to conventional unrolling, which often uses a predetermined number of layers and
may employ layer-specific parameters, SOCPNet uses a single shared update block that can be
iterated for a variable number of steps at inference~\citep{monga2021algorithm}. This
places our method closer to recurrent architectures and neural algorithmic reasoning, where
a shared computation is repeatedly applied to emulate an algorithmic
process~\citep{ibarz2022generalist}. We specialize this principle to IPM trajectories for
SOCP and study the training conditions required for stable long-horizon refinement. 
Two recent learned iterative solvers are related. PDHG-Net~\citep{pdhgnet2024} unrolls PDHG into a fixed-depth network with layer-specific parameters followed by a classical refinement step, and targets linear programs. In contrast, our setting is SOCP, our update is weight-shared across refinement steps, and the inference horizon is decoupled from the training horizon. IPM-LSTM~\citep{ipmlstm2024} is more closely related: it learns the inner operations of an IPM iteration for nonlinear-programming and warm-starts IPOPT with the result. By contrast, we learn a recursive update directly on the SOCP graph and introduce a stabilization procedure that enables practical long-horizon rollouts. In our experiments, SOCPNet outperforms an SOCP-adapted IPM-LSTM under identical conditions (Appendix~\ref{app:ipmlstm}). Finally, classical IPMs~\citep{wachter2006implementation,nocedal2006numerical} benefit from warm-starting~\citep{gould2000warm}, and differentiable conic solvers such as CALIPSO~\citep{howell2022calipso} still require a full solve per instance. We instead use the learned recursion to generate the warm start.


\noindent\textbf{Our approach.}
We emulate the IPM behavior of approaching the optimum through a sequence of progressively
refined iterates. To this end, we learn a single update rule and apply it recursively, so
that the network traces its own trajectory toward the solution. Sharing the same rule
across all steps allows the recursion to be evaluated for a variable number of iterations
at inference. Section~\ref{sec:method} describes the learned update rule.

\section{SOCPNet: A Recursive GNN for SOCP Solver Trajectories}
\label{sec:method}

SOCPNet learns a single per-iteration update
\begin{equation}
  \hat{\mathbf{x}}^{(t+1)} = \phi_\theta\!\left(\hat{\mathbf{x}}^{(t)}, u\right),
  \label{eq:update}
\end{equation}
with $u = (\mathbf{f}, \mathbf{F}, \mathbf{g}, \mathbf{l}, \mathbf{r},
\{\mathbf{A}_k, \mathbf{b}_k, \mathbf{c}_k, d_k\}_{k})$ the problem instance parameters
as defined in Eqs.~\eqref{eq:obj}-\eqref{eq:lin}. The parameters $\theta$ are shared across steps $t$, so
the same map is applied recursively for $T$ steps at inference. This
section specifies the two ingredients of the method, the network $\phi_\theta$ that realizes
one solver step (Section~\ref{sec:arch}) and the training procedure that keeps its recursion
accurate far beyond the supervision horizon (Section~\ref{sec:recipe}).

\subsection{Architecture: recursive message passing over the SOCP graph}
\label{sec:arch}

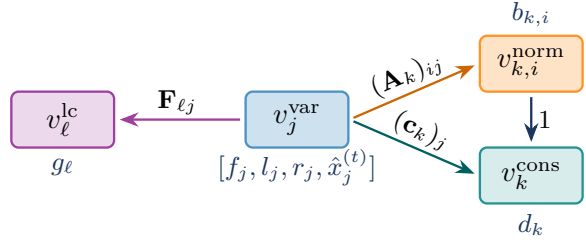
\begin{wrapfigure}{r}{0.55\textwidth}
\centering
\resizebox{\linewidth}{!}{%
\begin{tikzpicture}[
>=Stealth,
    vbox/.style={
        rounded corners=2.5pt,
        draw=VarBlue!80,
        fill=VarBlue!22,
        line width=0.8pt,
        minimum width=1.25cm,
        minimum height=6.8mm,
        inner xsep=4pt,
        inner ysep=1.5pt,
        align=center,
        font=\normalsize
    },
    nbox/.style={
        rounded corners=2.5pt,
        draw=orange!85,
        fill=orange!22,
        line width=0.8pt,
        minimum width=1.25cm,
        minimum height=6.8mm,
        inner xsep=4pt,
        inner ysep=1.5pt,
        align=center,
        font=\normalsize
    },
    cbox/.style={
        rounded corners=2.5pt,
        draw=ConsTeal!85,
        fill=ConsTeal!15,
        line width=0.8pt,
        minimum width=1.25cm,
        minimum height=6.8mm,
        inner xsep=4pt,
        inner ysep=1.5pt,
        align=center,
        font=\normalsize
    },
    lbox/.style={
        rounded corners=2.5pt,
        draw=violet!75,
        fill=violet!15,
        line width=0.8pt,
        minimum width=1.25cm,
        minimum height=6.8mm,
        inner xsep=4pt,
        inner ysep=1.5pt,
        align=center,
        font=\normalsize
    },
    st/.style={
        font=\normalsize\bfseries,
        color=PosterNavy
    },
    sub/.style={
        font=\small,
        color=PosterNavy,
        align=center
    },
    el/.style={
        font=\small,
        inner xsep=2pt,
        inner ysep=1.5pt,
        text=black
    },
    ar/.style={
        ->,
        line width=0.8pt,
        shorten >=1.5pt,
        shorten <=1.5pt
    }
]


\node[lbox] (l) at (-2.85,0.0) {$v_\ell^{\mathrm{lc}}$};

\node[vbox] (v) at (0.0,0.0) {$v_j^{\mathrm{var}}$};

\node[nbox] (n) at (2.85,0.68) {$v_{k,i}^{\mathrm{norm}}$};
\node[cbox] (c) at (2.85,-0.68) {$v_k^{\mathrm{cons}}$};

\node[sub] at (-2.85,-0.58) {$g_\ell$};
\node[sub] at (0.0,-0.58) {$[f_j,l_j,r_j,\hat{x}_j^{(t)}]$};
\node[sub] at (2.85,1.28) {$b_{k,i}$};
\node[sub] at (2.85,-1.28) {$d_k$};

\draw[ar, violet!75]
    (v.west) -- node[el, above, pos=0.52] {$\mathbf{F}_{\ell j}$} (l.east);

\draw[ar, orange!80!black]
    (v.east) -- node[el, above, sloped, pos=0.50] {$({\mathbf A}_k)_{ij}$} (n.west);

\draw[ar, ConsTeal!75!black]
    (v.east) -- node[el, above, sloped, pos=0.50] {$({\mathbf c}_k)_j$} (c.west);

\draw[ar, PosterNavy]
    (n.south) -- node[el, right, pos=0.50] {$1$} (c.north);

\end{tikzpicture}
}
\caption{Encoding of SOCP instance with parameters $u = (\mathbf{f}, \mathbf{F}, \mathbf{g}, \mathbf{l}, \mathbf{r},
\{\mathbf{A}_k, \mathbf{b}_k, \mathbf{c}_k, d_k\}_{k})$ into its graph.
}
\label{fig:socp2graph}
\end{wrapfigure}

\noindent\textbf{Graph and node/edge features.}
Given problem parameters
$u = (\mathbf{f}, \mathbf{F}, \mathbf{g}, \mathbf{l}, \mathbf{r},
\{\mathbf{A}_k, \mathbf{b}_k, \mathbf{c}_k, d_k\}_{k})$, we build a graph
$G=(V,E)$ with four node types that mirror the structure of the SOCP. Variable nodes
$v_j^{\mathrm{var}}$ have initial features
$[f_j,l_j,r_j,\hat{x}_j^{(t)}]$, consisting of the objective coefficient, lower and upper
bounds, and current iterate value. Norm-component nodes
$v_{k,i}^{\mathrm{norm}}$, one for each row $i$ of $\mathbf{A}_k$, have feature $b_{k,i}$.
Conic-constraint nodes $v_k^{\mathrm{cons}}$ have feature $d_k$, and linear-constraint
nodes $v_\ell^{\mathrm{lc}}$ have feature $g_\ell$.
Directed edges carry the corresponding problem coefficient as their edge feature:
variable-norm edges are weighted by $(\mathbf{A}_k)_{ij}$, variable-conic edges by
$(\mathbf{c}_k)_j$, variable-linear edges by $\mathbf{F}_{\ell j}$, and each norm node is
connected to its conic-constraint node, in both directions, with unit weight. The graph is a direct encoding of the constraint matrices, decomposing the nonlinear cone
constraint~\eqref{eq:cone} into linear interactions between typed nodes
(Fig.~\ref{fig:socp2graph}).

This graph is the representation of
\citet{li2025expressivity}. We adopt it together with their
message passing layers. In contrast to their end-to-end approach, which runs a single fixed-depth pass, we apply it recursively, feeding each prediction back as input.

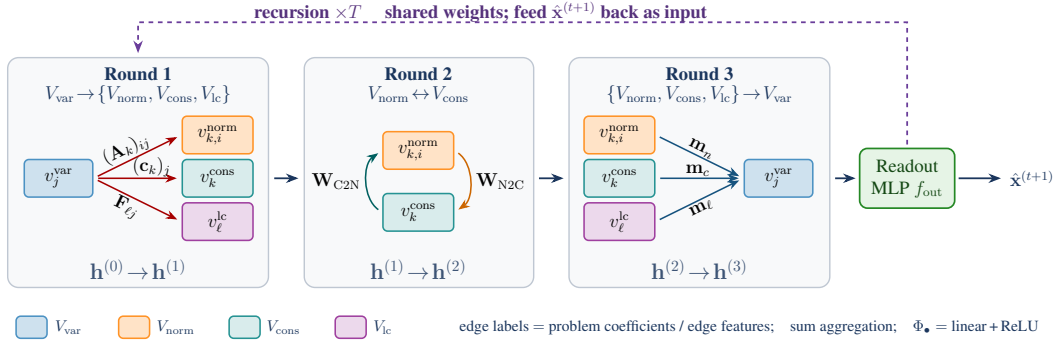
\begin{figure*}[t]
\centering
\resizebox{\linewidth}{!}{%
\begin{tikzpicture}[
    >=Stealth,
    vbox/.style={
        rounded corners=3pt,
        draw=VarBlue!80,
        fill=VarBlue!22,
        line width=1pt,
        minimum width=1.55cm,
        minimum height=8.5mm,
        inner xsep=4pt,
        inner ysep=2pt,
        align=center,
        font=\large
    },
    nbox/.style={
        rounded corners=3pt,
        draw=orange!85,
        fill=orange!22,
        line width=1pt,
        minimum width=1.60cm,
        minimum height=8.5mm,
        inner xsep=4pt,
        inner ysep=2pt,
        align=center,
        font=\large
        },
        cbox/.style={
            rounded corners=3pt,
            draw=ConsTeal!85,
            fill=ConsTeal!15,
            line width=1pt,
            minimum width=1.60cm,
            minimum height=8.5mm,
            inner xsep=4pt,
            inner ysep=2pt,
            align=center,
            font=\large
    },
    lbox/.style={
        rounded corners=3pt,
        draw=violet!75,
        fill=violet!15,
        line width=1pt,
        minimum width=1.60cm,
        minimum height=8.5mm,
        inner xsep=4pt,
        inner ysep=2pt,
        align=center,
        font=\large
    },
    panel/.style={
        draw=PosterNavy!28,
        line width=0.9pt,
        rounded corners=7pt,
        fill=PosterNavy!3
    },
    st/.style={
        font=\large\bfseries,
        color=PosterNavy
    },
    sub/.style={
        font=\large,
        color=PosterNavy,
        align=center
    },
    state/.style={
        font=\Large\bfseries,
        color=PosterNavy!85
    },
    el/.style={
        font=\large,
        inner xsep=2pt,
        inner ysep=1pt,
        text=black
    },
    ar/.style={
        ->,
        line width=1pt,
        shorten >=3pt,
        shorten <=3pt
    },
    big/.style={
        ->,
        line width=1.15pt,
        color=PosterNavy,
        shorten >=2pt,
        shorten <=2pt
    },
    rdo/.style={
        rounded corners=5pt,
        draw=TrajGreen!85!black,
        fill=TrajGreen!12,
        line width=1.1pt,
        minimum width=2.15cm,
        minimum height=1.35cm,
        align=center,
        font=\large,
        text=TrajGreen!45!black
    },
    legendtext/.style={
        font=\normalsize,
        color=PosterNavy
    }
]

\draw[panel] (0.00,-2.65) rectangle (5.90,2.55);

\node[st]  at (2.95,2.18) {Round 1};
\node[sub] at (2.95,1.72)
    {$V_{\mathrm{var}}\!\to\!\{V_{\mathrm{norm}},V_{\mathrm{cons}},V_{\mathrm{lc}}\}$};

\node[vbox] (v1) at (1.15,-0.18) {$v_j^{\mathrm{var}}$};

\node[nbox] (n1) at (4.75,0.80) {$v_{k,i}^{\mathrm{norm}}$};
\node[cbox] (c1) at (4.75,-0.18) {$v_k^{\mathrm{cons}}$};
\node[lbox] (l1) at (4.75,-1.16) {$v_\ell^{\mathrm{lc}}$};

\draw[ar, red!65!black]
    (v1.east) -- node[el, above, sloped, pos=0.43] {$({\mathbf A}_k)_{ij}$} (n1.west);

\draw[ar, red!65!black]
    (v1.east) -- node[el, above, pos=0.66] {$({\mathbf c}_k)_j$} (c1.west);

\draw[ar, red!65!black]
    (v1.east) -- node[el, below, sloped, pos=0.43] {$\mathbf F_{\ell j}$} (l1.west);

\node[state] at (2.95,-2.23) {$\mathbf h^{(0)}\!\to\!\mathbf h^{(1)}$};

\draw[panel] (6.71,-2.65) rectangle (11.90,2.55);

\node[st]  at (9.29,2.18) {Round 2};
\node[sub] at (9.29,1.72)
    {$V_{\mathrm{norm}}\!\leftrightarrow\!V_{\mathrm{cons}}$};

\node[nbox] (n2) at (9.29,0.45) {$v_{k,i}^{\mathrm{norm}}$};
\node[cbox] (c2) at (9.29,-0.95) {$v_k^{\mathrm{cons}}$};

\draw[ar, orange!80!black]
    (n2.east) to[out=-35,in=35]
    node[el, right=2pt, pos=0.50] {$\mathbf W_{\mathrm{N2C}}$}
    (c2.east);

\draw[ar, ConsTeal!75!black]
    (c2.west) to[out=145,in=-145]
    node[el, left=2pt, pos=0.50] {$\mathbf W_{\mathrm{C2N}}$}
    (n2.west);

\node[state] at (9.29,-2.23) {$\mathbf h^{(1)}\!\to\!\mathbf h^{(2)}$};

\draw[panel] (12.70,-2.65) rectangle (18.60,2.55);

\node[st]  at (15.65,2.18) {Round 3};
\node[sub] at (15.65,1.72)
    {$\{V_{\mathrm{norm}},V_{\mathrm{cons}},V_{\mathrm{lc}}\}\!\to\!V_{\mathrm{var}}$};

\node[nbox] (n3) at (13.84,0.80) {$v_{k,i}^{\mathrm{norm}}$};
\node[cbox] (c3) at (13.84,-0.18) {$v_k^{\mathrm{cons}}$};
\node[lbox] (l3) at (13.84,-1.16) {$v_\ell^{\mathrm{lc}}$};

\node[vbox] (v3) at (17.44,-0.18) {$v_j^{\mathrm{var}}$};

\draw[ar, VarBlue!70!black]
    (n3.east) -- node[el, above, sloped, pos=0.48] {$\mathbf m_n$} (v3.west);

\draw[ar, VarBlue!70!black]
    (c3.east) -- node[el, above, pos=0.48] {$\mathbf m_c$} (v3.west);

\draw[ar, VarBlue!70!black]
    (l3.east) -- node[el, below, sloped, pos=0.48] {$\mathbf m_\ell$} (v3.west);

\node[state] at (15.65,-2.23) {$\mathbf h^{(2)}\!\to\!\mathbf h^{(3)}$};

\node[rdo] (rd) at (20.35,-0.18)
    {Readout\\MLP $f_{\mathrm{out}}$};

\draw[big] (18.65,-0.18) -- (rd.west);

\draw[big] (rd.east) -- ++(1.10,0)
    node[right, font=\large, color=PosterNavy] {$\hat{\mathbf x}^{(t+1)}$};

\draw[big] (5.96,-0.18) -- (6.65,-0.18);
\draw[big] (11.95,-0.18) -- (12.65,-0.18);

\draw[big, TrajPurple, dashed]
    (rd.north) |- (10.46,3.35) -| (2.95,2.58);

\node[
    font=\large\bfseries,
    color=TrajPurple!80!black
] at (10.70,3.58)
    {recursion $\times T$ \quad shared weights; feed $\hat{\mathbf{x}}^{(t+1)}$ back as input};

\node[vbox, minimum width=7mm, minimum height=5mm, inner sep=0pt] at (0.54,-3.55) {};
\node[legendtext, right] at (0.95,-3.55) {$V_{\mathrm{var}}$};

\node[nbox, minimum width=7mm, minimum height=5mm, inner sep=0pt] at (2.86,-3.55) {};
\node[legendtext, right] at (3.25,-3.55) {$V_{\mathrm{norm}}$};

\node[cbox, minimum width=7mm, minimum height=5mm, inner sep=0pt] at (5.36,-3.55) {};
\node[legendtext, right] at (5.74,-3.55) {$V_{\mathrm{cons}}$};

\node[lbox, minimum width=7mm, minimum height=5mm, inner sep=0pt] at (7.76,-3.55) {};
\node[legendtext, right] at (8.15,-3.55) {$V_{\mathrm{lc}}$};

\node[
    legendtext,
    right,
    align=left
] at (10.10,-3.55)
    {edge labels $=$ problem coefficients / edge features;\quad sum aggregation;\quad
     $\Phi_\bullet=$ linear\,+\,ReLU};
\end{tikzpicture}
}%
\caption{SOCPNet's steps: three-round heterogeneous message
passing (Eqs.~\eqref{eq:mp1}--\eqref{eq:mp3}).}
\label{fig:mp_scheme}
\end{figure*}

\noindent\textbf{Message passing.}
Let $\mathbf{h}_a^{(\ell)}\in\R^d$ denote the embedding of node $a$ after message
passing round $\ell$ within recursive step $t$. The initial embedding
$\mathbf{h}_a^{(0)}$ is obtained by applying a learned linear map to the corresponding node
features. Each round aggregates type-specific messages using an edge-type-specific weight
matrix $\mathbf{W}_\bullet\in\R^{d\times d}$. Messages are summed over neighbors and
weighted by the scalar edge feature. The resulting incoming message is concatenated with the
previous node embedding and passed through a linear-plus-ReLU update
$\Phi_\bullet$. Each recursive step performs three message passing rounds,
$\mathbf{h}_a^{(0)}\!\to\!\mathbf{h}_a^{(1)}\!\to\!\mathbf{h}_a^{(2)}\!\to\!\mathbf{h}_a^{(3)}$
(Fig.~\ref{fig:mp_scheme}).

\noindent\emph{Round 1}
(\textcolor{VarBlue!70!black}{\textbf{variables}} $\to$
\textcolor{orange!80!black}{\textbf{norm}}/
\textcolor{ConsTeal!70!black}{\textbf{conic}}/
\textcolor{violet!70!black}{\textbf{linear}})
sends the current variable information to all constraint-related nodes. For a norm node,
and analogously for conic and linear nodes,
\begin{align}
  \mathbf{m}_{v_{k,i}^{\mathrm{norm}}}^{(1)} &= \!\!\!\sum_{v_j^{\mathrm{var}}\in\mathcal N(v_{k,i}^{\mathrm{norm}})}\!\!\!(\mathbf{A}_k)_{ij}\,\mathbf{W}_{\mathrm{V2N}}\,\mathbf{h}_{v_j^{\mathrm{var}}}^{(0)}, &
  \mathbf{h}_{v_{k,i}^{\mathrm{norm}}}^{(1)} &= \Phi_1\!\big([\mathbf{h}_{v_{k,i}^{\mathrm{norm}}}^{(0)},\mathbf{m}_{v_{k,i}^{\mathrm{norm}}}^{(1)}]\big).
  \label{eq:mp1}
\end{align}
\noindent\emph{Round 2}
(\textcolor{orange!80!black}{\textbf{norm}} $\leftrightarrow$
\textcolor{ConsTeal!70!black}{\textbf{conic}})
exchanges information between each cone and its norm components:
\begin{align}
  \mathbf{h}_{v_k^{\mathrm{cons}}}^{(2)} &= \Phi_2\!\big([\mathbf{h}_{v_k^{\mathrm{cons}}}^{(1)}, \textstyle\sum_{i}\mathbf{W}_{\mathrm{N2C}}\mathbf{h}_{v_{k,i}^{\mathrm{norm}}}^{(1)}]\big), &
  \mathbf{h}_{v_{k,i}^{\mathrm{norm}}}^{(2)} &= \Phi_3\!\big([\mathbf{h}_{v_{k,i}^{\mathrm{norm}}}^{(1)}, \mathbf{W}_{\mathrm{C2N}}\mathbf{h}_{v_k^{\mathrm{cons}}}^{(2)}]\big).
  \label{eq:mp2}
\end{align}
\noindent\emph{Round 3}
(\textcolor{orange!80!black}{\textbf{norm}}/
\textcolor{ConsTeal!70!black}{\textbf{conic}}/
\textcolor{violet!70!black}{\textbf{linear}}
$\to$ \textcolor{VarBlue!70!black}{\textbf{variables}})
gathers the constraint information back onto each variable node:
\begin{equation}
  \mathbf{h}_{v_j^{\mathrm{var}}}^{(3)} = \Phi_4\!\big([\mathbf{h}_{v_j^{\mathrm{var}}}^{(0)},\, \mathbf{m}_c,\, \mathbf{m}_n,\, \mathbf{m}_\ell]\big),
  \label{eq:mp3}
\end{equation}
where $\mathbf{m}_c,\mathbf{m}_n,\mathbf{m}_\ell$ are the edge-weighted sums of
$\mathbf{W}_{\mathrm{C2V}}\mathbf{h}^{(2)}$, $\mathbf{W}_{\mathrm{N2V}}\mathbf{h}^{(2)}$,
and $\mathbf{W}_{\mathrm{LC2V}}\mathbf{h}^{(1)}$ over the conic, norm, and linear
neighbors of $v_j^{\mathrm{var}}$, respectively. These three rounds constitute one learned
solver step. Because they define $\phi_\theta$ and their parameters are shared across
recursive steps, repeated application of this block realizes the recursion in
Eq.~\eqref{eq:update}.

\noindent\textbf{Readout to a trajectory prediction.}
A two-layer MLP $f_{\mathrm{out}}$ maps the final variable embedding
$\mathbf{h}_{v_j^{\mathrm{var}}}^{(3)}$, together with four graph-level context vectors,
to the next iterate:
\begin{equation}
  \hat{x}_j^{(t+1)} = f_{\mathrm{out}}\!\big(
    \mathbf{h}_{v_j^{\mathrm{var}}}^{(3)},\;
    \mathbf{I}_{\mathrm{var}},\;
    \mathbf{I}_{\mathrm{norm}},\;
    \mathbf{I}_{\mathrm{cons}},\;
    \mathbf{I}_{\mathrm{lc}}\big),
  \label{eq:readout}
\end{equation}
where $\mathbf{I}_\bullet=\sum_{a\in V_\bullet}\mathbf{h}_a$ denotes the sum of embeddings
over all nodes of type $\bullet$. Thus, each variable readout depends both on its local
embedding and on a global summary of the instance. A shared readout maps the final embeddings to $\hat{\mathbf{x}}^{(t+1)}$, fed back as input; the same weights run all $T\gg T_{\mathrm{train}}$ steps (purple feedback in Fig.~\ref{fig:mp_scheme}).

Equation~\eqref{eq:readout} defines the base architecture. Two design choices remain: how to parametrize the update and what information each variable node
should receive. We therefore treat the corresponding modifications as variants of the same
network. These variants leave the message passing scheme in
Eqs.~\eqref{eq:mp1}--\eqref{eq:mp3} unchanged and modify only the readout or the input
features. They are motivated in Section~\ref{sec:recipe} and evaluated 
in the ablation study of Section~\ref{sec:ablation}.

In the \emph{step-size control} variant, the readout emits a per-graph log-magnitude and a
per-node unit direction for the update
$\hat{\mathbf{x}}^{(t+1)}-\hat{\mathbf{x}}^{(t)}$. In the \emph{KKT-informed} variant, the
variable feature $\hat{x}_j^{(t)}$ is augmented with local cone, linear, and box slacks
computed at the current iterate before Round~1. Thus, each variable node also receives
information about how close its associated constraints are to becoming active or violated.
In all cases, applying the readout at every recursive step yields the predicted trajectory
$\{\hat{\mathbf{x}}^{(1)},\dots,\hat{\mathbf{x}}^{(T)}\}$.

For the next iteration, variable-node features are updated by concatenating the
prediction $\hat{\mathbf{x}}^{(t)}$ with the final embedding
$\mathbf{h}_{v_j^{\mathrm{var}}}^{(3)}$ through a learnable linear-plus-ReLU layer, which
becomes the initial embedding for iteration $t+1$, realizing the shared-weight recursion
of \eqref{eq:update}.

\subsection{Making the recursion work: architecture and loss ingredients}
\label{sec:recipe}

We construct SOCPNet incrementally. We start from the simplest version of the recursion in
Eq.~\eqref{eq:update}, namely a recursive GNN trained by direct trajectory matching, and
then add architectural and loss components one at a time to improve accuracy and reduce
long-horizon drift. This section introduces each component in order. The ablation in Section~\ref{sec:ablation} adds these components sequentially (Table~\ref{tab:ablation-design}, steps S1 to S9), so each step in the table corresponds to a paragraph below.

\paragraph{Data and unrolling.}
A training sample consists of an SOCP instance $u$ and the IPOPT trajectory it generates,
$\mathbf{x}^{(0)},\dots,\mathbf{x}^{(T_u)}$, whose final iterate is treated as the reference
solution $\mathbf{x}^\star$. Its prediction $\hat{\mathbf{x}}^{(t)}$ feeds the next step. It is supervised at
\emph{every} step $t$, i.e., each readout $\hat{\mathbf{x}}^{(t)}$ is scored against the
corresponding solver iterate $\mathbf{x}^{(t)}$ (trajectory supervision) or against the
optimum $\mathbf{x}^\star$ (target supervision). Weight sharing lets the same map unroll
for $T\gg T_{\text{train}}$ steps at inference. Because solver trajectories have different lengths, we pad each
trajectory to a common horizon $T$ by repeating $\mathbf{x}^\star$. By default, these padded
steps are masked out of the losses below; $\operatorname{mean}_t$ denotes the resulting
average over supervised, unmasked steps. For each sample, the recursion is initialized at
the solver's starting point, $\hat{\mathbf{x}}^{(0)}=\mathbf{x}^{(0)}$, unrolled for $T$
steps on the graph of $u$, and trained by backpropagation through the full unroll. Thus, the
shared parameters $\theta$ receive gradient contributions from every supervised step. All
losses are also averaged over the instances in each mini-batch.

\paragraph{Trajectory matching (S3).}
The most direct training objective is to ask the prediction at step $t$ to reproduce the
corresponding solver iterate (Fig.~\ref{fig:trajlearn}):
\begin{equation}
  \mathcal{L}_{\mathrm{mse}}
  = \operatorname{mean}_t\, \|\hat{\mathbf{x}}^{(t)}-\mathbf{x}^{(t)}\|_2^2 .
  \label{eq:lmse}
\end{equation}
This objective aligns the learned recursion with the solver trajectory step by step.

\paragraph{Residual updates (S4): predicting a correction, not a position.}
In the base readout of Eq.~\eqref{eq:readout}, the network predicts the next iterate
directly, $\hat{\mathbf{x}}^{(t+1)}=f_{\mathrm{out}}(\cdot)$. Early in training, this can
cause each step to overwrite the current iterate with an arbitrary output. We therefore
parametrize the readout as a correction $\Delta_\theta$ added to the current iterate:
\begin{equation}
  \hat{\mathbf{x}}^{(t+1)}
  = \hat{\mathbf{x}}^{(t)} + \Delta_\theta\big(\hat{\mathbf{x}}^{(t)}, u\big).
  \label{eq:residual}
\end{equation}
This residual form stabilizes the recursion: when
$\Delta_\theta\approx\mathbf{0}$, the iterate remains close to its current value rather than
moving arbitrarily. All recursive variants from this point onward use
Eq.~\eqref{eq:residual}.

\paragraph{Soft-DTW (S5): matching the path, not the pace.}
A learned update need not move in increments of the same size as an interior-point method.
We therefore replace the MSE main loss with soft-DTW~\citep{cuturi2017soft}, which compares
the predicted and reference trajectories up to a differentiable monotone realignment
$\pi$ over the admissible set $\mathcal{A}$:
\begin{equation}
  \mathcal{L}_{\mathrm{dtw}}
  = {\min}^{\gamma}_{\pi\in\mathcal{A}}\ \sum_{(i,j)\in\pi}
    \|\hat{\mathbf{x}}^{(i)}-\mathbf{x}^{(j)}\|_2^2 ,
  \qquad
  {\min}^{\gamma}\{a_i\} = -\gamma\log\textstyle\sum_i e^{-a_i/\gamma},
  \label{eq:ldtw}
\end{equation}
with $\gamma=0.1$. Soft-DTW encourages the learned recursion to follow the solver path while
allowing it to progress along that path at a different rate.

\paragraph{Why trajectory matching drifts.}
In our experiments, both Eq.~\eqref{eq:lmse} and Eq.~\eqref{eq:ldtw} produce recursions that
reach $\sre\approx10^{-2}$ within a few steps but then drift when unrolled beyond the
training horizon: the iterates stop improving and move away from the solution, causing the
error to grow with additional steps (Fig.~\ref{fig:ablation-small}). Because these losses
are based on squared errors, their gradients become very small once the residual reaches
approximately $10^{-2}$, creating a precision floor in the regime where further accuracy is
needed. Moreover, neither loss explicitly penalizes distance to the terminal solution
$\mathbf{x}^\star$, so the learned update is not discouraged from moving away from the
optimum; under repeated application, such deviations can accumulate.

\paragraph{Target supervision on a log scale (S6).}
We replace the trajectory-matching loss with one that scores every prediction  $\hat{\mathbf{x}}^{(t)}$ against the optimum $\mathbf{x}^\star$ instead of the solver's iterate $\mathbf{x}^{(t)}$,
\begin{equation}
  \sre_t^2 = \frac{\|\hat{\mathbf{x}}^{(t)}-\mathbf{x}^\star\|_2^2}
                  {\max\!\big(1,\|\mathbf{x}^\star\|_2^2\big)},
  \qquad
  \mathcal{L}_{\mathrm{tgt}}
  = \operatorname{mean}_t\, \tfrac{1}{2}\log\!\big(\sre_t^2+\varepsilon\big) ,
  \qquad \varepsilon = 10^{-12}.
  \label{eq:sre}
\end{equation}
The normalization makes instances of very different magnitude comparable, and the logarithm
weights a relative gain equally at every scale, so an improvement from $10^{-1}$ to $10^{-2}$
counts as much as one from $10^{-4}$ to $10^{-5}$. The trajectory still shapes training through $\mathbf{x}^{(0)}$, the horizon, and the padded tail, but the network is now free to reach $\mathbf{x}^\star$ along its own path.

\noindent With a stable recursion (S4) and a scale-free target loss (S6) in place, we add the
remaining ingredients together (S7): three penalties on the shape of the sequence, a
step-size parametrization of the update, KKT-informed input features, and objective and
constraint supervision.

\paragraph{Shaping the sequence: contraction, anchor, fixed point.}
The target loss constrains the value of each iterate but not the shape of the trajectory
between iterates. We add three penalties for that. Since contraction is a property of consecutive steps rather than single iterates, a \emph{Picard penalty} enforces step-wise distance reduction to the optimum by a factor $\rho$~\citep{banach1922operations}. An \emph{anchor} fixes the last supervised iterate, while a \emph{fixed-point} term guarantees that $\mathbf{x}^\star$ remains unchanged once reached.
\begin{equation}
  \mathcal{L}_{\mathrm{pic}} = \operatorname{mean}_t\, \mathrm{relu}\big(d_{t+1}-\rho\,d_t\big) ,
  \qquad
  \mathcal{L}_{\mathrm{anc}} = \sre_{t_{\mathrm{last}}}^2 ,
  \qquad
  \mathcal{L}_{\mathrm{fp}} = \ \|\phi_\theta(\mathbf{x}^\star,u)-\mathbf{x}^\star\|_2^2 ,
  \label{eq:picard}
\end{equation}
with $d_t=\sre_t$ the relative error at step $t$, $\rho=0.9$, and $t_{\mathrm{last}}$ the last
supervised step. The fixed-point term is the only place $\mathbf{x}^\star$ is fed to the
network as an \emph{input}.

\paragraph{Step-size control.}
The residual $\Delta_\theta$ of \eqref{eq:residual} is a single vector responsible for two
decisions at once: the direction of the move and its length. We let the readout emit two
quantities instead, a per-node unit direction $\hat{\mathbf{d}}_t\in\R^n$ with
$\|\hat{\mathbf{d}}_t\|_2=1$ and a single per-graph log-magnitude $\hat{s}_t\in\R$, and
recombine them into the correction,
\begin{equation}
  \Delta_\theta = e^{\hat{s}_t}\,\hat{\mathbf{d}}_t ,
  \qquad
  \hat{\mathbf{d}}_t = \frac{\mathbf{z}_t}{\|\mathbf{z}_t\|_2} ,
  \label{eq:polar}
\end{equation}
where $\mathbf{z}_t$ is the raw per-node output. Since a solver takes large steps early and
very small ones near the optimum, spanning several orders of magnitude, we parametrize the
length in the log domain and supervise it against the solver's own step length,
\begin{equation}
  \mathcal{L}_{\mathrm{step}} = \operatorname{mean}_t\, (\hat{s}_t-s_t^\star)^2 ,
  \qquad
  s_t^\star = \log\|\mathbf{x}^{(t+1)}-\mathbf{x}^{(t)}\|_2 .
  \label{eq:step}
\end{equation}
This leaves one interpretable scalar per graph, on which the network can learn to slow down
as it approaches the solution instead of overshooting it.

\paragraph{KKT-informed features: tracking feasibility.}
At an optimum, the objective gradient is balanced by the active constraints and every
constraint is either slack or exactly tight; these are the Karush-Kuhn-Tucker
conditions~\citep{nocedal2006numerical}. The per-constraint slacks are the
residuals an interior-point method monitors on its way there, so we give each variable node
the slacks of the current iterate as extra input features,
\begin{equation}
  s^{\mathrm{soc}}_k = \mathbf{c}_k^\top\hat{\mathbf{x}}^{(t)}+d_k
    - \|\mathbf{A}_k\hat{\mathbf{x}}^{(t)}+\mathbf{b}_k\|_2, \quad
  s^{\mathrm{lin}}_\ell = g_\ell - \mathbf{F}_\ell\hat{\mathbf{x}}^{(t)}, \quad
  s^{\mathrm{box}}_j = \big(\hat{x}^{(t)}_j-l_j,\; r_j-\hat{x}^{(t)}_j\big).
  \label{eq:slacks}
\end{equation}
Each is positive exactly when its constraint holds, and its magnitude says how much room is
left, so the next update can depend on how close to, and on which side of, feasibility the
iterate lies. The raw coordinates carry the same information only implicitly.

\paragraph{Objective and constraint supervision.}
A solution is judged not only by its distance to $\mathbf{x}^\star$ but also by its objective
value and its feasibility, so we supervise both along the trajectory:
$\mathcal{L}_{\mathrm{obj}}$ is the squared gap to $\mathbf{f}^\top\mathbf{x}^\star$,
normalized by $\max(1,(\mathbf{f}^\top\mathbf{x}^\star)^2)$, and $\mathcal{L}_{\mathrm{cons}}$
is the squared cone, linear, and box violations. This completes S7: the network is now
trained on a weighted sum of $\mathcal{L}_{\mathrm{tgt}}$, $\mathcal{L}_{\mathrm{fp}}$,
$\mathcal{L}_{\mathrm{pic}}$, $\mathcal{L}_{\mathrm{anc}}$, $\mathcal{L}_{\mathrm{step}}$,
$\mathcal{L}_{\mathrm{obj}}$, and $\mathcal{L}_{\mathrm{cons}}$ (Eq.~\eqref{eq:total} in
Appendix~\ref{app:training}).

\paragraph{Tail supervision (S8).}
By default the padded steps beyond the solver's own length $T_u$ are masked out, so the
network is supervised only where the solver produced an iterate. Full tail supervision scores
those padded steps as well, against $\mathbf{x}^\star$, so the losses reward staying at the
optimum for the entire unrolled horizon rather than only up to $T_u$. This directly targets
the long-horizon behavior the recursion is actually run for at inference, penalizing any
late drift away from $\mathbf{x}^\star$.

\paragraph{Capacity (S9).}
Finally, we widen the node embedding from $d{=}64$ to $d{=}256$. Capacity is added last, once
the recursion is already stable, so the extra parameters refine an update that already
converges rather than fitting a drifting one. In the experiments, we use the SOCPNet model with $d{=}256$.

\paragraph{Inference-time projection.}
At inference, we add a short projection, not used during training: a few projected-gradient
steps on the total constraint violation push $\hat{\mathbf{x}}$ toward feasibility while
leaving $\sre$ and the objective essentially unchanged. Training details are
given in Appendix~\ref{app:training}.

\section{Experiments}
\label{sec:exp}

Our analysis is organized around four questions. First, does \emph{supervision} matter? In other words,
is a recursive model trained on the solver's whole trajectory more accurate than a network
regressing only its endpoint? Second, can our model remain stable beyond its supervision
horizon? Third, does our method \emph{generalize} across SOCP families and to larger
problems, with one fixed set of hyperparameters? Fourth, is the learned solution
useful as a \emph{warm start} for an exact solver? We first describe the datasets, protocol and metrics, then answer each question.

\begin{table}[t]
  \centering
  \setlength{\tabcolsep}{4pt}
  
  \caption{Unstructured datasets test results.}
  \label{tab:main}
  \resizebox{.9\textwidth}{!}{
  \begin{tabular}{llcccc}
    \toprule
    Dataset & Model & $\sre$ & absObj & relObj & Viol \\
    \midrule
    \multirow{6}{*}{Small ($n{=}50$)}
      & Direct MLP                        & $2.5\times10^{-1}$ & $3.0\times10^{0\phantom{-}}$ & $3.2\times10^{-1}$ & $5.9\times10^{-3}$ \\
      & SOCP-GNN ($d{=}64$)                       & $2.9\times10^{-3}$ & $7.4\times10^{-2}$ & $7.7\times10^{-3}$ & $1.3\times10^{-3}$ \\
      & SOCP-GNN ($d{=}256$)                      & $3.3\times10^{-3}$ & $2.9\times10^{-1}$ & $3.1\times10^{-2}$ & $8.8\times10^{-4}$ \\
      & SOCPNet ($d{=}64$)               & $2.1\times10^{-5}$ & $9.8\times10^{-3}$ & $1.1\times10^{-3}$ & $2.3\times10^{-4}$ \\
      & SOCPNet ($d{=}256$)              & $\mathbf{3.1\times10^{-6}}$ & $\mathbf{4.6\times10^{-3}}$ & $\mathbf{5.7\times10^{-4}}$ & $\mathbf{1.3\times10^{-4}}$ \\
      & IPOPT (reference)                 & $\sim 0$ & $\sim 10^{-4}$ & $\sim 10^{-5}$ & $\sim 10^{-9}$ \\
    \midrule
    \multirow{6}{*}{Medium ($n{=}100$)}
      & Direct MLP                        & $7.5\times10^{-1}$ & $1.6\times10^{1\phantom{-}}$ & $8.7\times10^{-1}$ & $4.2\times10^{-2}$ \\
      & SOCP-GNN ($d{=}64$)                  & $1.3\times10^{-2}$ & $2.4\times10^{-1}$ & $1.4\times10^{-2}$ & $2.9\times10^{-3}$ \\
      & SOCP-GNN ($d{=}256$)                 & $1.3\times10^{-2}$ & $4.1\times10^{-1}$ & $2.3\times10^{-2}$ & $1.9\times10^{-3}$ \\
      & SOCPNet ($d{=}64$)               & $1.2\times10^{-3}$ & $\mathbf{2.2\times10^{-2}}$ & $\mathbf{1.2\times10^{-3}}$ & $\mathbf{1.9\times10^{-4}}$ \\
      & SOCPNet ($d{=}256$)              & $\mathbf{9.6\times10^{-4}}$ & $2.7\times10^{-2}$ & $1.4\times10^{-3}$ & $\mathbf{1.9\times10^{-4}}$ \\
      & IPOPT (reference)                 & $\sim 0$ & $\sim 10^{-4}$ & $\sim 10^{-5}$ & $\sim 10^{-9}$ \\
    \bottomrule
  \end{tabular}
  }
\end{table}

\subsection{Experimental setup: datasets, protocol, and metrics}
\label{sec:setup}

\noindent\textbf{Datasets.} We consider four families of datasets, each with two different graph sizes, \emph{Small} ($n{=}50$, $m{=}10$ conic, $b{=}10$ linear constraints) and \emph{Medium} ($n{=}100$, $m{=}50$, $b{=}50$).
The first family, named \emph{Unstructured}, is composed of SOCP instances randomly generated with the procedure described in~\citet{li2025expressivity}.
In addition to this benchmark, we also build three structured families, whose full
SOCP formulations are given in Appendix~\ref{app:families}. \emph{Robust ellipsoid} minimizes a linear objective whose coefficients are only known to lie in an ellipsoidal uncertainty set, so the constraint becomes a norm ball and the problem a
second-order cone program; this is the canonical construction of robust linear optimization~\citep{bental1999robust,lobo1998applications}. \emph{Block angular} programs consist of otherwise independent blocks tied together by a small number of linking constraints, the structure exploited by decomposition methods and typical of large-scale
resource-allocation and stochastic programs~\citep{dantzig1960decomposition,birge2011introduction}.
\emph{Portfolio risk} bounds a quadratic risk term, written as a second-order cone constraint, while maximizing expected return, the standard conic form of mean-variance portfolio selection~\citep{markowitz1952portfolio,lobo1998applications}.
Each dataset contains $5{,}000$ instances with a 80/10/10 split for training/validation/testing.

\noindent\textbf{Protocol and metrics.} All instances are solved with IPOPT to obtain ground-truth trajectories (tolerance $10^{-5}$). Our model is trained over $T_{\text{train}}=26$ unrolling steps on the Small datasets and $T_{\text{train}}=21$ steps on the Medium datasets.
For testing, we compare three learned models against IPOPT. Direct MLP flattens the problem data and infers the optimum $\mathbf{x}^\star$ in one forward pass. SOCP-GNN is the end-to-end model proposed by~\citet{li2025expressivity}, which processes the graph of Fig.~\ref{fig:socp2graph} and predicts $\mathbf{x}^\star$ in a single shot. SOCPNet is our recursive model. They are both  reported at two capacities $d{=}64$ and $d{=}256$. The number of unrolling iterations for testing is greater than $T_{\text{train}}$, i.e.,  respectively $h{=}30$ and $h{=}25$ for Small and Medium. This way we can evaluate the model's accuracy beyond its training horizon.
For each of the above models, we report the solution relative error $\sre$ (Eq.~\eqref{eq:sre}), the absolute/relative objective error at median (absObj/relObj), and the mean constraint violation before post-hoc feasibility restoration (Viol) at each iterate. In Table~\ref{tab:main}, the best learning solution per column, per dataset block is in bold.

\subsection{Does recursion help?}
\label{sec:target}

\noindent\textbf{Recursion outperforms direct prediction.} Table~\ref{tab:main}
compares the two single-shot baselines with SOCPNet. On Small, SOCPNet reaches
$\sre = 3.1\times10^{-6}$, several orders of magnitude below  SOCP-GNN
($2.9\times10^{-3}$) and the Direct MLP ($2.5\times10^{-1}$), and the same ordering holds on
Medium. This advantage depends on the training procedure of Section~\ref{sec:recipe}, as the
ablation shows.

In Appendix~\ref{app:compute-matched}, we further compare SOCPNet against a compute-matched SOCP-GNN that repeats the single-shot message passing block $25$ times, matching SOCPNet's inference complexity but without iterate feedback. On Small, this deeper one-shot model performs similarly to the shallow SOCP-GNN with $\sre$ of about $3.3\times10^{-3}$, whereas SOCPNet
achieves $\sre=3.1\times10^{-6}$. This indicates that the gain is due to recursive refinement rather than additional message passing alone.

\subsection{Ablation study: which ingredients make the recursion work?}
\label{sec:ablation}
\begin{wraptable}[15]{r}{0.5\textwidth}
    \vspace{-15pt}
    \centering
    \small
    \caption{Ablation study and baselines.}
    \label{tab:ablation-design}
    \setlength{\tabcolsep}{3pt}
    \renewcommand{\arraystretch}{1.05}
    \begin{tabularx}{\linewidth}{
        @{}
        >{\centering\arraybackslash}p{0.42cm}
        >{\centering\arraybackslash}p{0.55cm}
        >{\raggedright\arraybackslash}X
        @{}
    }
        \toprule
        & \textbf{Step} & \textbf{Description} \\
        \midrule
        \multirow{2}{*}{\rotatebox{90}{\scriptsize Base}}
        & S1 & MLP (flatten $\to$ MLP, single-shot, MSE) \\
        & S2 & SOCP-GNN (single-shot, MSE) \\
        \midrule
        \multirow{6}{*}{%
            \rotatebox{90}{\scriptsize Ablated SOCPNet}%
        }
        & S3 & Recursive GNN, MSE (base SOCPNet) \\
        & S4 & + residual updates \\
        & S5 & Soft-DTW loss (replaces MSE) \\
        & S6 & Log-SRE loss (replaces Soft-DTW) \\
        & S7 & + fixed-point/Picard/anchor, step-size control,
               objective/constraint losses, KKT features \\
        & S8 & + full tail supervision \\
        \midrule
        \rotatebox{90}{\scriptsize Full}
        & S9 & \textbf{SOCPNet} (width $d=64 \rightarrow 256$) \\
        \bottomrule
    \end{tabularx}
\end{wraptable}
In Table~\ref{tab:ablation-design}, S1-S2 are single-shot baselines; S3-S8 are ablated variants of SOCPNet, each adding or replacing one component on top of the row above; S9 is the full model. Steps are cumulative.

\begin{figure}[t]
  \centering
  \includegraphics[trim={0 0.2cm 0 0.3cm},clip, width=\linewidth]{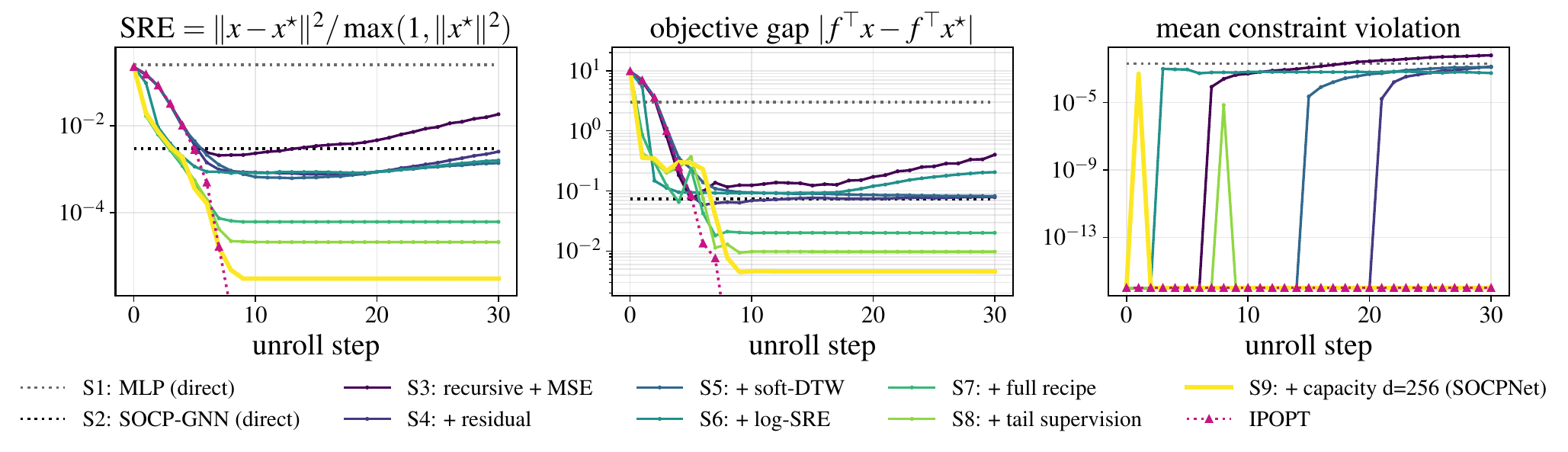}
  \caption{Unstructured Small benchmark, nine-step ablation
  (Table~\ref{tab:ablation-design}). Median SRE, median objective gap, and
  mean constraint violation vs.\ unroll step ($\downarrow$ better; log axes).}
  \label{fig:ablation-small}
\end{figure}

\noindent\textbf{Design.} We run a controlled  ablation on the Unstructured Small and Medium datasets, adding one ingredient per step and holding everything else fixed (e.g., same splits, same optimizer). Each step defines one training setting labeled  in order of their introduction.
S3 is the base version of SOCPNet and S9 is the complete procedure. The ingredients are listed in Table~\ref{tab:ablation-design}, and we refer to the settings by these labels throughout. The ablation results in terms of median SRE, median objective gap, and mean constraint violation are shown in Fig.~\ref{fig:ablation-small}  for Small. The Medium curves show the same trends and are deferred to Appendix~\ref{app:ablation-medium}, Fig.~\ref{fig:ablation-medium}.

\noindent\textbf{Main findings.} We summarize here the three main findings. \textit{(i)} \emph{Recursion alone is not enough.} A plain recursive GNN trained by trajectory MSE (S3) is no better than the single-shot GNN (S2), and its error starts growing once it is unrolled past its training horizon. \textit{(ii)} \emph{The training procedure makes recursion effective.} Adding residual
connections (S4), replacing MSE with Soft-DTW trajectory matching (S5), then supervising against the optimum rather than the solver's iterates (S6), and adding the penalties (S7) lowers the error by orders of magnitude and flattens the curves: the upward drift disappears, and the recursion keeps improving well
beyond the horizon it was trained on. \textit{(iii)} \emph{Capacity pays off last.}
Supervising the padded tail (S8) and widening the embedding (S9) give the final gains, but
only once the recursion is stable. S9 achieves a median SRE of about $3\times10^{-6}$ in Unstructured Small and $\sim9.6\times10^{-4}$ in Unstructured Medium. S9 is the complete model we call SOCPNet.
 
\noindent\textbf{Additional findings.}
Constraint violation is driven down to near machine precision ($\sim\!10^{-15}$) by the
post-hoc feasibility pass for every variant, so the accuracy gains do not impact feasibility.


\noindent\textbf{Trajectory dependence.}
IPOPT trajectories are used in the S9 model only to supervise the step-size control of Eq.~\eqref{eq:polar}. To isolate this remaining trajectory dependence, we add an ablation study to Appendix~\ref{app:stepsize} that keeps the full SOCPNet design unchanged but removes the solver-supervised step-size target. On Small, the median SRE remains comparable to the full model ($4.8\times 10^{-7}$ versus $3.1\times 10^{-6}$), while the mean SRE increases from $1.4\times 10^{-3}$ to $2.2\times 10^{-3}$. Thus, trajectory supervision through the step-size target mainly improves stability by reducing high-error rollouts.

\subsection{Experiments on other SOCP families}
\label{sec:generalization}

The results so far establish SOCPNet on the Unstructured benchmark. Under identical data and protocol, we train both SOCPNet and SOCP-GNN on the three families of Section~\ref{sec:setup}. 

\begin{figure}[tb]
  \centering
  \includegraphics[trim={0 0.2cm 0 0.25cm},clip, width=\linewidth]{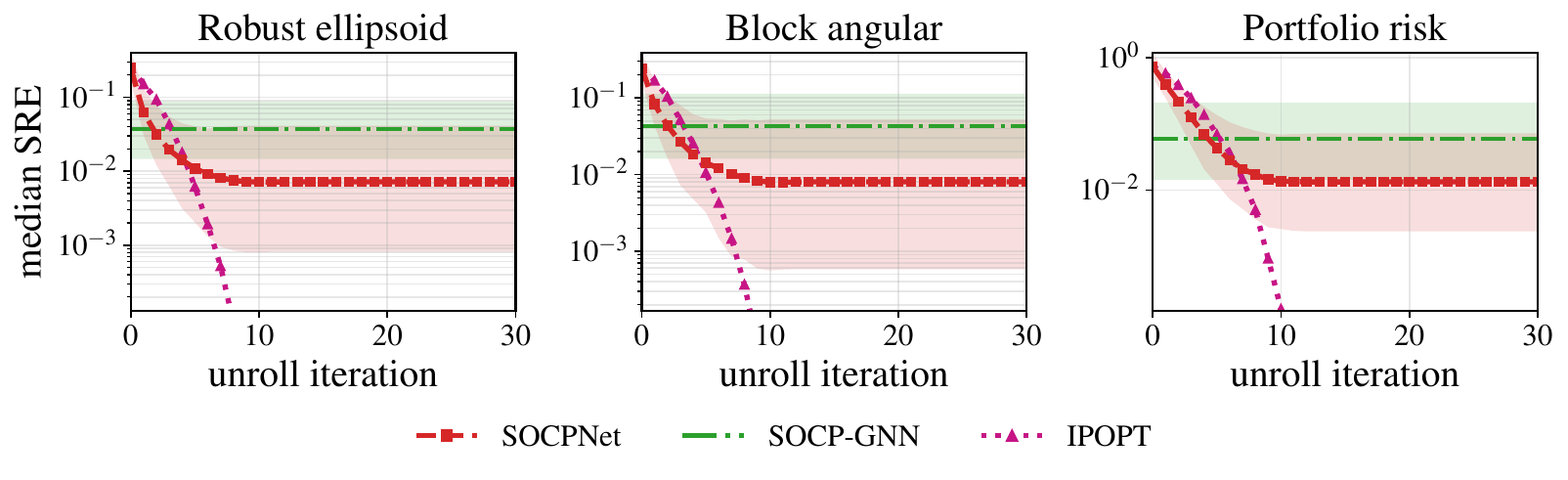}
  \caption{Small Structured families: SOCPNet vs.\ SOCP-GNN vs.\ IPOPT.
  Per-step median SRE ($\downarrow$ better; log axis). Bands are $95\%$ percentile intervals.}
  \label{fig:generalization}
\end{figure}

In Fig.~\ref{fig:generalization} we can observe that SOCPNet converges without drift on every family and reaches a median SRE $4-5\times$ lower than SOCP-GNN, so the gain is not specific to the Unstructured benchmark the procedure was developed on. The same ordering holds with the Medium datasets on robust ellipsoid and block angular,
where SOCPNet again reaches lower error than the SOCP-GNN; on portfolio risk the two are
comparable and SOCPNet does not improve on the baseline at Medium scale. Full per-family
curves are deferred to Appendix~\ref{app:medium-families}.

\subsection{Hybrid solver: Warm-starting a classical solver with SOCPNet}
\label{sec:hybrid}

\begin{figure}[tb]
  \centering
  \includegraphics[width=\linewidth]{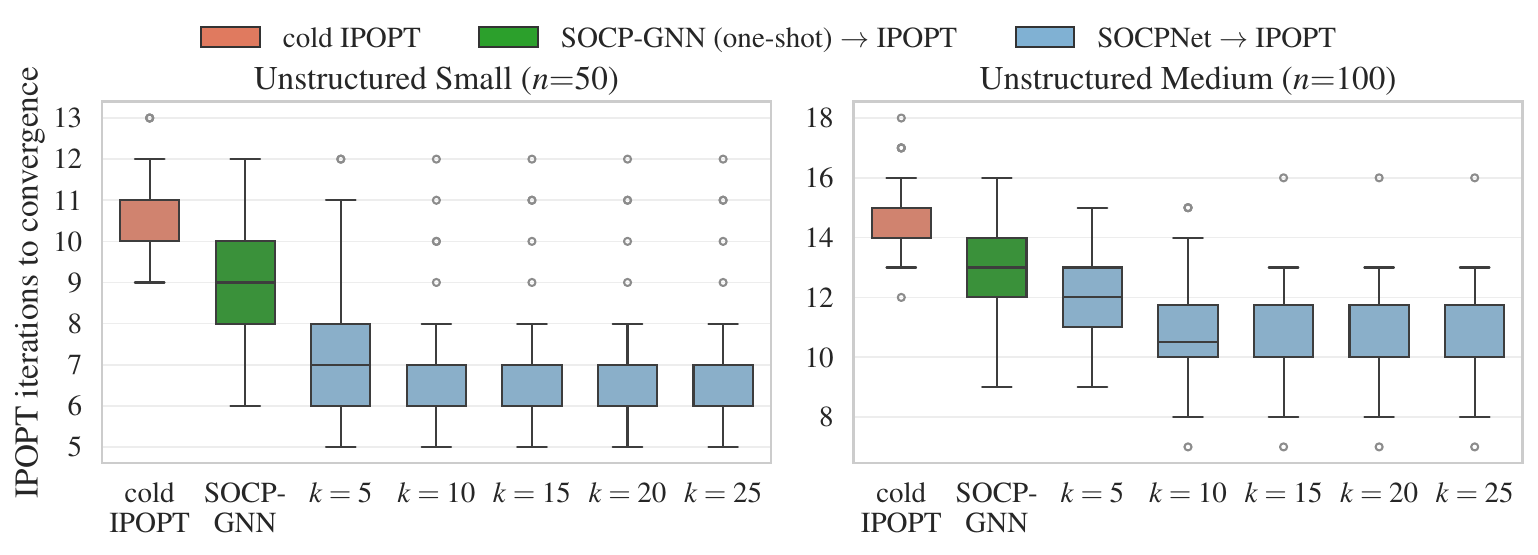}
  \caption{Hybrid solver: IPOPT iterations to convergence ($\downarrow$ better).
  Cold start, one-shot SOCP-GNN warm start, and SOCPNet warm start ($d{=}256$)
  at handoff step $k$, on the Unstructured
  Small (left) and Medium (right) datasets ($50$ test problems each).}
  \label{fig:hybrid}
\end{figure}

As a data-driven method, SOCPNet is not guaranteed to find the optimal solution $\mathbf{x}^\star$, which is desired in many applications. Here, we propose a hybrid approach that uses SOCPNet to obtain a fast approximation of $\mathbf{x}^\star$, then hands off to a classical solver for exactness. We run the trained recursion for $k$ steps and use $\hat{\mathbf{x}}^{(k)}$ as the initial point for IPOPT, leaving all solver internals unchanged.
We sweep $k \in \{0,5,10,15,20,25\}$, where
$k=0$ denotes the cold-start baseline. As seen in Fig.~\ref{fig:hybrid}, a few SOCPNet steps cut
the median iteration count from $11$ to $6$ in Small and from $15$ to $10$ in Medium,
plateauing at $k{\geq}10$. Final accuracy is unaffected, see Appendix~\ref{app:medium-hybrid}, Figure~\ref{fig:hybrid-accuracy}.

\noindent To highlight the benefit of recursion, we also warm-start IPOPT with the SOCP-GNN baseline, shown as the second box in Fig.~\ref{fig:hybrid}. Cold-start IPOPT requires a median of $11$/$15$ iterations on Small/Medium, the one-shot SOCP-GNN warm start reduces this to $9$/$13$, and the SOCPNet warm start further reduces it to
$6$/$10$. Thus, the one-shot predictor provides a modest reduction of $18\%$/$13\%$, whereas the
recursive SOCPNet warm start nearly halves the IPOPT iteration count.

On Small the saved iterations translate directly into end-to-end time, cutting the median runtime from $17.0$\,s to $8.0$\,s at $k{=}15$ (a $2.1\times$ speed-up, largest over $k\in[10,20]$). On Medium the same reduction gives no speed-up, because a fixed per-call IPOPT overhead dominates. The full breakdown is in Appendix~\ref{app:medium-hybrid}, Table~\ref{tab:hybrid-time}. We therefore
report iteration count as the primary hybrid metric.

\section{Discussion and Conclusion}
\label{sec:discussion}

Interior-point methods already expose a rich sequence of intermediate states, and our results show that these states can be used to learn a more accurate SOCP solver than direct endpoint regression. However, recursion alone is not sufficient: a shared GNN trained with standard trajectory matching drifts beyond the training horizon. Stable refinement only emerges once the update is explicitly controlled, with step-size control playing a decisive role, together with residual updates, log-SRE target supervision, KKT-informed features, objective and constraint losses, and tail supervision.
The learned iterates are also useful for warm-starting a classical solver. When used to initialize a classical solver, namely IPOPT, they reduce its iteration count by $45\%$ on Small and $33\%$ on Medium, with no loss in final accuracy. These gains are not uniform across all settings: on Medium portfolio-risk instances, SOCPNet does not improve over the baseline. A key direction for future work is therefore to identify the problem characteristics that enable stable trajectory learning across different SOCP families and scales.

\section*{Acknowledgments}
We would like to thank
R\'emy Flamary, Professor at \'Ecole Polytechnique, and the anonymous reviewers for their valuable discussions and insights.

{\small
\setlength{\bibsep}{1.5pt plus 0.3ex}
\bibliographystyle{unsrtnat}
\bibliography{reference_revised}
}

\appendix

\section*{Appendix}
This appendix is organized as follows. Appendix~\ref{app:AIPolicy} states how generative AI
tools were used in this work. Appendix~\ref{app:families} gives the formulations and the
instance generation of the SOCP problem families, and Appendix~\ref{app:training} details the
training setup, hyperparameters, loss definitions, and inference-time projection.
Appendix~\ref{app:ablation-medium} reports the nine-step ablation on the Unstructured Medium
benchmark. Appendix~\ref{app:compute-matched} compares SOCPNet with a compute-matched
single-shot SOCP-GNN, and Appendix~\ref{app:ipmlstm} compares it with an SOCP adaptation of
IPM-LSTM. Appendix~\ref{app:stepsize} isolates the effect of the solver-derived step-size
signal. Appendix~\ref{app:medium-families} repeats the comparison on the structured families at
Medium scale, Appendix~\ref{app:zero-shot} evaluates zero-shot transfer across problem scales,
and Appendix~\ref{app:loo} evaluates leave-one-out generalization across problem families.
Appendix~\ref{app:medium-hybrid} reports the final solution accuracy and the wall-clock timing
of the hybrid warm-start solver. Finally, Appendix~\ref{app:runtable} describes the resources
provided to reproduce our experiments.

\section{Generative AI Usage Statement}
\label{app:AIPolicy}
In this work, we used generative AI tools (large language models) for language editing and for assistance with writing and refining the manuscript's prose, for basic coding assistance in producing figures and analysis scripts, and for drafting portions of the text that were subsequently revised by the authors. All core research content, including the method, the training procedure, the experimental design, and the analysis of results, was conceived and carried out by the authors. We reviewed all AI-assisted work: AI-assisted text was checked and edited by the authors, AI-assisted code was verified and tested for correctness, and every reference was checked against its primary source to confirm it exists and is cited accurately. We take responsibility for the final content of this work, including all text, claims, and artifacts produced with the aid of generative AI.

\section{SOCP Problem Families}
\label{app:families}

Every dataset is an instance of the standard SOCP of Eqs.~\eqref{eq:obj}--\eqref{eq:lin},
\begin{equation}
  \min_{\mathbf{x}\in\R^n}\ \mathbf{f}^\top\mathbf{x}
  \quad\text{s.t.}\quad
  \|\mathbf{A}_k\mathbf{x}+\mathbf{b}_k\|_2 \le \mathbf{c}_k^\top\mathbf{x}+d_k
  \ \ (k\in[m]),
  \qquad
  \mathbf{F}\mathbf{x}\le\mathbf{g},
  \qquad
  \mathbf{l}\le\mathbf{x}\le\mathbf{r} .
  \label{eq:fam-socp}
\end{equation}
The families differ only in how the data
$(\mathbf{f},\{\mathbf{A}_k,\mathbf{b}_k,\mathbf{c}_k,d_k\},\mathbf{F},\mathbf{g},
\mathbf{l},\mathbf{r})$ are generated. Problem sizes $(n,m,b)$ take the Small and Medium
values of Section~\ref{sec:exp}, and the Unstructured benchmark follows the random-SOCP generator
described in~\citet{li2025expressivity}.

\paragraph{Feasibility by construction.}
Sampling constraint data at random usually yields an infeasible or unbounded program, so we
build each structured instance around a point that is feasible by construction. We first draw
a reference point $\mathbf{x}_{\mathrm{ref}}\sim\mathcal{N}(0,\mathbf{I})$ and sample the box
$[\mathbf{l},\mathbf{r}]$ so that it brackets $\mathbf{x}_{\mathrm{ref}}$. The constraint
data $\mathbf{F}$ and $\{\mathbf{A}_k,\mathbf{b}_k,\mathbf{c}_k\}$ are then sampled as sparse
Gaussians, and only the right-hand sides are chosen, so that every constraint holds at
$\mathbf{x}_{\mathrm{ref}}$ with a margin:
\begin{equation}
  \mathbf{g} = \mathbf{F}\mathbf{x}_{\mathrm{ref}} + \boldsymbol{\zeta} ,
  \qquad
  d_k = \|\mathbf{A}_k\mathbf{x}_{\mathrm{ref}}+\mathbf{b}_k\|_2
        - \mathbf{c}_k^\top\mathbf{x}_{\mathrm{ref}} + \epsilon_k ,
  \label{eq:fam-feasible}
\end{equation}
with slacks $\boldsymbol{\zeta}\ge\mathbf{0}$ and small margins $\epsilon_k>0$. By
construction, $\mathbf{x}_{\mathrm{ref}}$ is feasible, so the instance is solvable; it is not
in general optimal, and plays no role beyond generation. Instances that IPOPT fails to solve
to tolerance $10^{-5}$ are discarded.

\paragraph{Robust ellipsoid.}
A linear program whose constraint coefficients are uncertain, known only to lie in an
ellipsoid, admits a robust counterpart with one cone constraint per uncertain row: asking the
row to hold for every realization in the ellipsoid is equivalent to
\begin{equation}
  \|\mathbf{A}_k\mathbf{x}+\mathbf{b}_k\|_2
  \ \le\ -\mathbf{a}_k^\top\mathbf{x}+d_k ,
  \qquad k\in[m] .
  \label{eq:fam-robust}
\end{equation}
The left side is the worst-case deviation over the uncertainty set, and the right side the
nominal margin left by $\mathbf{x}$; in the notation of \eqref{eq:fam-socp} this is simply
$\mathbf{c}_k=-\mathbf{a}_k$. We sample $\mathbf{f}\sim\mathcal{N}(0,\tfrac14\mathbf{I})$
and, per cone, a sparse Gaussian matrix $\mathbf{A}_k$, a sparse vector $\mathbf{b}_k$, and a
sparse Gaussian vector $\mathbf{a}_k$.

\paragraph{Block angular.}
Here the variables form groups that would be independent were it not for a few constraints
that couple them. We partition the $n$ coordinates into disjoint blocks
$\mathcal{B}_1,\dots,\mathcal{B}_B$, with $B$ drawn uniformly from $\{4,\dots,10\}$, and let
each cone act on a single block,
\begin{equation}
  \|\mathbf{A}_k\mathbf{x}_{\mathcal{B}_{p(k)}}+\mathbf{b}_k\|_2
  \ \le\ \mathbf{c}_k^\top\mathbf{x}_{\mathcal{B}_{p(k)}}+d_k ,
  \qquad
  \mathbf{F}\mathbf{x}\le\mathbf{g} ,
  \label{eq:fam-block}
\end{equation}
where $p(k)$ denotes the block assigned to cone $k$, so $\mathbf{A}_k$ is nonzero only on
that block's coordinates. The linear rows $\mathbf{F}\mathbf{x}\le\mathbf{g}$ are the only
constraints spanning several blocks. Dropping them would leave $B$ independent subproblems;
keeping them yields the classical block-angular structure of large-scale
resource-allocation and stochastic programs, where many nearly separable subproblems are tied
together by a small number of linking rows.

\paragraph{Portfolio risk.}
A portfolio $\mathbf{x}$ maximizes expected return subject to a bound on its risk. Risk is
the standard deviation $\sqrt{\mathbf{x}^\top\boldsymbol{\Sigma}\mathbf{x}}$ under an asset
covariance $\boldsymbol{\Sigma}$, and factoring
$\boldsymbol{\Sigma}=\mathbf{R}^\top\mathbf{R}$ turns that bound into a cone constraint,
since $\sqrt{\mathbf{x}^\top\boldsymbol{\Sigma}\mathbf{x}}=\|\mathbf{R}\mathbf{x}\|_2$. We
impose one such bound per asset subset $S_k$, together with a budget row,
\begin{equation}
  \|\mathbf{R}_k\mathbf{x}_{S_k}+\mathbf{b}_k\|_2 \ \le\ d_k ,
  \qquad
  \mathbf{w}^\top\mathbf{x}\le g_0 ,
  \qquad
  \mathbf{w}\ge\mathbf{0},\ \ \textstyle\sum_j w_j = 1 ,
  \label{eq:fam-portfolio}
\end{equation}
which is \eqref{eq:fam-socp} with $\mathbf{A}_k=\mathbf{R}_k$ on the columns $S_k$ and
$\mathbf{c}_k=\mathbf{0}$. To sample an instance we form
$\boldsymbol{\Sigma}=\mathbf{M}\mathbf{M}^\top+0.06\,\mathbf{I}$ from a Gaussian $\mathbf{M}$,
the diagonal addition keeping $\boldsymbol{\Sigma}$ positive definite, and take $\mathbf{R}_k$
as the Cholesky factor of the principal submatrix $\boldsymbol{\Sigma}_{S_k}$, with a small
$\mathbf{b}_k$. The objective uses $\mathbf{f}\le\mathbf{0}$, so minimizing
$\mathbf{f}^\top\mathbf{x}$ maximizes expected return. This is the conic form of mean-variance portfolio selection.

\section{Training Setup and Hyperparameters}
\label{app:training}

All models use AdamW with gradient clipping at $1.0$ and a fixed random seed ($42$).
Baseline steps use a constant learning rate $5\times10^{-4}$. The full procedure (S7 onward) uses a cosine schedule with $5$ warmup epochs. Batch size is $16$ throughout the ablation. The $d{=}256$ full model uses batch size $16$ (Small). Training horizons are $T_{\text{train}}{=}26$ for Small (padded) and $T_{\text{train}}{=}21$ for Medium.
Training budgets range from $80$ epochs to $200$ epochs depending on the model and problem size. Embedding dimension is $d{=}64$ through the procedure
build-up (S3-S8) and $d{=}256$ for the full model (S9). Models are implemented in PyTorch
and PyTorch Geometric and trained on NVIDIA RTX A4000 GPUs. IPOPT ground truth is obtained via
CyIpopt with tolerance $10^{-5}$ and a $10$-minute per-instance timeout. Direct baselines, S1 and S2, are trained end-to-end against $\mathbf{x}^\star$ with MSE.

\noindent\textbf{Epoch budgets and checkpoint selection.} The epoch budget depends on the
model. The single-shot baselines
(Direct MLP and SOCP-GNN) output one prediction and converge quickly, so they run for $80$
(Small) and $100$ (Medium) epochs. The recursive SOCPNet has to predict an entire trajectory
and trains for up to $200$ epochs. For every training run, we monitor the validation loss at each epoch. We keep the best-validation checkpoint for evaluation. Under this policy, the validation loss of all runs has plateaued before their budget is exhausted. Thus, all models have converged. Batch size, optimizer, gradient clipping, and seed are held fixed across every run.

\noindent\textbf{Loss definitions.} The complete training objective is
\begin{equation}
  \mathcal{L} = \mathcal{L}_{\mathrm{tgt}}
  + \lambda_{\mathrm{fp}}\,\mathcal{L}_{\mathrm{fp}}
  + \lambda_{\mathrm{pic}}\,\mathcal{L}_{\mathrm{pic}}
  + \lambda_{\mathrm{anc}}\,\mathcal{L}_{\mathrm{anc}}
  + \lambda_{\mathrm{step}}\,\mathcal{L}_{\mathrm{step}}
  + \lambda_{\mathrm{obj}}\,\mathcal{L}_{\mathrm{obj}}
  + \lambda_{\mathrm{cons}}\,\mathcal{L}_{\mathrm{cons}} ,
  \label{eq:total}
\end{equation}

with weights chosen through grid search on the Unstructured benchmark's validation split,
then fixed once and reused unchanged across every dataset and problem family we report.

All losses are averaged over the batch and, where
indicated, over the $T$ unrolled steps. $\hat{\mathbf{x}}^{(t)}$ is the prediction and
$\mathbf{x}^{\star,(t)}$ the IPOPT ground-truth iterate (padded with $\mathbf{x}^\star$
beyond IPOPT's own length, for tail supervision). 

\noindent\textbf{Inference-time projection.} At inference only (not during training), we
take $10$ to $40$ projected-gradient steps on the total constraint violation, updating
$\mathbf{x} \leftarrow \mathbf{x} - \mathrm{clip}\big(\alpha\nabla_{\mathbf{x}}\tfrac{1}{2}
\|\mathrm{relu}(\mathrm{viol}(\mathbf{x}))\|^2\big)$ with $\alpha=1.0$ and per-coordinate
clipping between $0.03$ and $0.05$. On the Small benchmark (Section~\ref{sec:setup}), this
lowers the mean constraint violation from $1.3\times10^{-4}$ to about $6\times10^{-10}$
while leaving $\sre$ and the objective essentially unchanged.

\section{Medium-scale Ablation Curves}
\label{app:ablation-medium}

\begin{figure}[t]
  \centering
  \includegraphics[trim={0 0.2cm 0 0.3cm},clip, width=\linewidth]{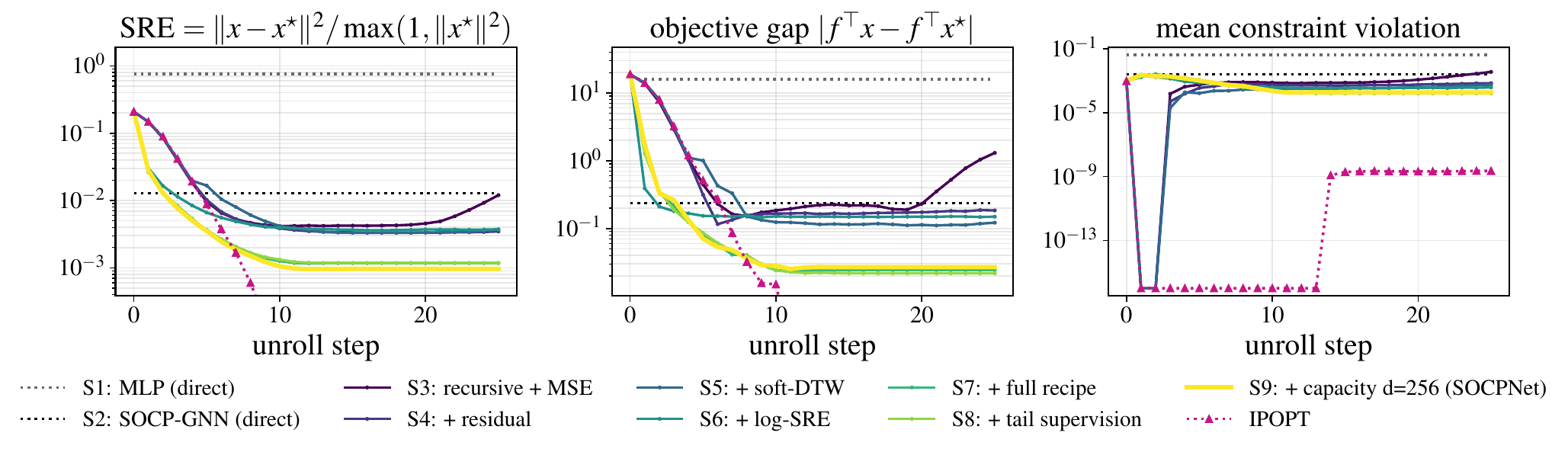}
  \caption{Unstructured Medium benchmark, nine-step ablation. Same conventions and labels as Fig.~\ref{fig:ablation-small}.}
  \label{fig:ablation-medium}
\end{figure}

Figure~\ref{fig:ablation-medium} repeats the nine-step ablation of
Section~\ref{sec:ablation} on the Unstructured Medium benchmark. The ordering of the steps and
the qualitative behavior are the same as at Small scale: the trajectory-matching variants drift
past the training horizon, the stabilized variants do not, and the error floor is reached only
after S7.

\section{Compute-matched Comparison}
\label{app:compute-matched}

\begin{figure}[t]
  \centering
  \includegraphics[width=0.45\linewidth]{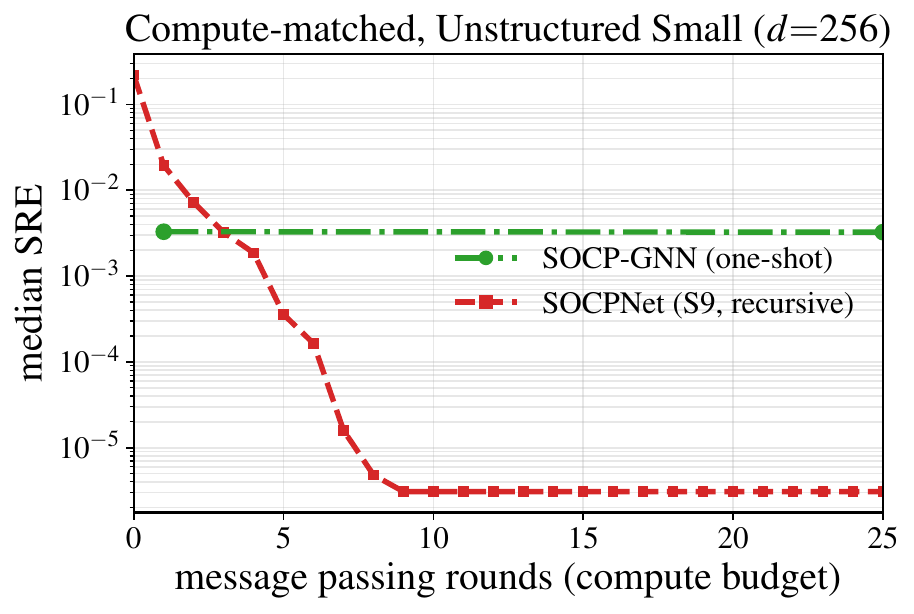}
\caption{Compute-matched comparison on Unstructured Small ($d{=}256$). Median $\sre$
  ($\downarrow$ better; log axis) against the number of message passing block applications.
  SOCP-GNN is plotted as a constant line because it is single-shot, but it uses the same budget
  of $25$ message passing rounds as SOCPNet at the rightmost point. Repeating its block $25$
  times does not improve accuracy, whereas SOCPNet descends over that budget.}

  \label{fig:compute-matched}
\end{figure}

SOCPNet applies its message passing block once per recursive step, so an unrolled SOCPNet
performs many times the message passing of a single-shot SOCP-GNN. To check that its advantage is
not simply a consequence of that larger budget, we train a single-shot SOCP-GNN ($d{=}256$) whose
message passing block is repeated $25$ times inside one forward pass. This matches the number of
block applications SOCPNet performs over the horizon reported in Fig.~\ref{fig:compute-matched}, with no feedback of the
predicted iterate. The two models then differ only in the recursion; everything else follows
the protocol of Section~\ref{sec:setup}.

On Unstructured Small, in Fig.~\ref{fig:compute-matched}, the compute-matched model reaches a median $\sre$ of
$3.25\times10^{-3}$, indistinguishable from the shallow single-shot SOCP-GNN
($3.29\times10^{-3}$) and three orders of magnitude above SOCPNet ($3.08\times10^{-6}$) at the
same budget. On Unstructured Medium, the
compute-matched model reaches a median $\sre$ of $6.8\times10^{-2}$,
worse than the shallow single-shot SOCP-GNN at this scale ($1.3\times10^{-2}$, Table~\ref{tab:main})
and about $70\times$ above SOCPNet ($9.6\times10^{-4}$). Thus confirming the results again on the larger dataset.

\section{Comparison with IPM-LSTM}
\label{app:ipmlstm}

IPM-LSTM~\citep{ipmlstm2024} is a shared-weight recurrent
network trained to accelerate IPM. We compare against it under a matched and fair protocol. We built an SOCP
adaptation of IPM-LSTM, which was originally validated only on QPs/QCQPs, and
compared it with SOCPNet under identical conditions.

\begin{table}[h!]
  \centering
  \caption{Median $\sre$ on the Small Unstructured test set ($500$ instances, $h{=}30$)
  under identical training and evaluation. IPM-LSTM adapted to SOCP against SOCPNet, the
  single-shot SOCP-GNN, and the IPOPT reference ($\downarrow$ better).}
  \label{tab:ipmlstm}
  \small
  \begin{tabular}{lccccc}
    \toprule
    Step & IPOPT & SOCPNet ($d{=}256$) & SOCPNet ($d{=}64$) & SOCP-GNN & IPM-LSTM \\
    \midrule
    $25$ & $\sim 0$ & $3.1\times10^{-6}$ & $2.1\times10^{-5}$ & $3.3\times10^{-3}$ & $2.0\times10^{-2}$ \\
    \bottomrule
\end{tabular}
\end{table}

\begin{figure}[h!]
  \centering
  \includegraphics[width=0.5\linewidth]{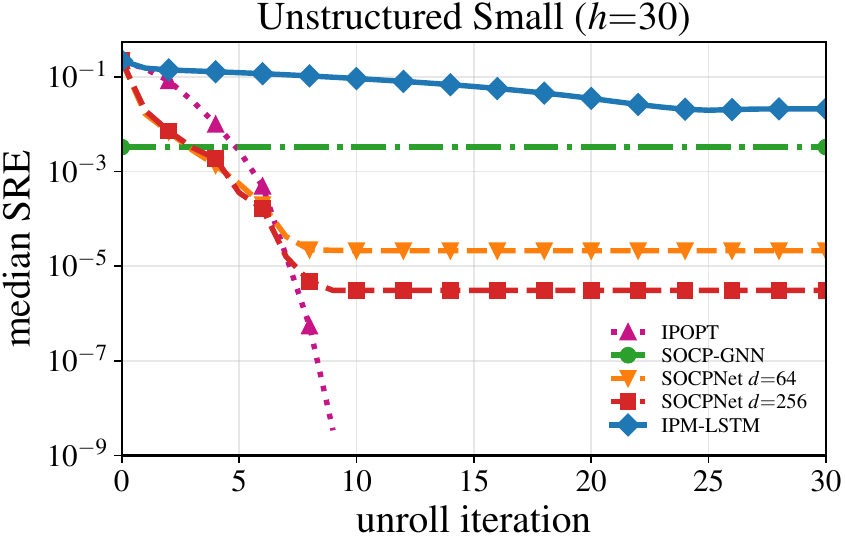}
  \caption{Median normalized $\sre$ over unroll steps on the Small Unstructured test set
  ($500$ instances, $h{=}30$; $\downarrow$ better, log axis). IPM-LSTM peaks near step $25$ and
  then drifts, while both SOCPNet variants ($d{=}64$ and $d{=}256$) converge and hold orders of
  magnitude below IPM-LSTM.}
  \label{fig:ipmlstm_overlay}
\end{figure}

\begin{table}[h!]
  \centering
  \caption{Median $\sre$ on the Medium Unstructured test set ($500$ instances)
  under identical training and evaluation ($\downarrow$ better).}
  \label{tab:ipmlstm_medium}
  \small
  \begin{tabular}{lccccc}
    \toprule
    Step & IPOPT & SOCPNet ($d{=}256$) & SOCPNet ($d{=}64$) & SOCP-GNN & IPM-LSTM \\
    \midrule
    $21$ & $\sim 0$ & $9.6\times10^{-4}$ & $1.2\times10^{-3}$ & $1.3\times10^{-2}$ & $3.7\times10^{-2}$ \\
    \bottomrule
  \end{tabular}
\end{table}

\begin{figure}[h!]
  \centering
  \includegraphics[width=0.5\linewidth]{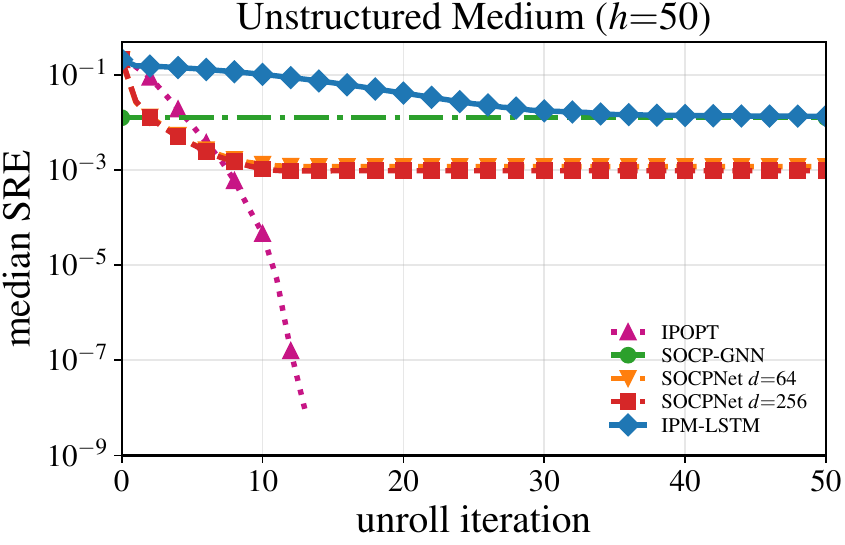}
  \caption{Median normalized $\sre$ over unroll steps on the Medium Unstructured test set
  ($500$ instances, $h{=}30$; $\downarrow$ better, log axis). Both SOCPNet variants remain
  about an order of magnitude below IPM-LSTM}
  \label{fig:ipmlstm_overlay_medium}
\end{figure}

Each cone constraint
\begin{equation*}
\|\mathbf{A}_k\mathbf{x}+\mathbf{b}_k\|_2 - \mathbf{c}_k^\top\mathbf{x} - d_k \le 0
\end{equation*}
is supplied as a smooth, twice-differentiable inequality, providing the residual, gradient, and
Hessian contributions that IPM-LSTM is trained to invert at every interior-point iteration.
Each cone residual is evaluated through $\|\mathbf{z}\|_2=\sqrt{\mathbf{z}^\top\mathbf{z}}$, which is not differentiable at $\mathbf{z}=\mathbf{A}_k\mathbf{x}+\mathbf{b}_k=\mathbf{0}$. To keep the gradient and Hessian used by IPM-LSTM's linear-system model well-defined everywhere, we regularize the norm as $\|\mathbf{z}\|_2\approx\sqrt{\mathbf{z}^\top\mathbf{z}+\varepsilon}$ with $\varepsilon{=}10^{-8}$, which is standard practice.
The classical IPM loop is otherwise unchanged. We use the same Small Unstructured split as SOCPNet. Relative to the published IPM-LSTM configuration (QP/QCQP benchmarks in \citet{ipmlstm2024}), we changed only the outer IPM horizon, from $T_{\mathrm{outer}}{=}100$ to $T_{\mathrm{outer}}{=}25$, to match SOCPNet's training horizon. The dataset size also differs: $10{,}000$ instances in their work and $5{,}000$ in our SOCP dataset. All other hyperparameters are kept at their published values: inner LSTM unroll $T_{\mathrm{inner}}{=}50$, hidden size $50$, learning rate $10^{-4}$, KKT preconditioning enabled, centering parameter $\sigma{=}0.9$, and fraction-to-the-boundary parameter $\tau{=}0.999$. We trained for up to $200$ epochs with early stopping. Both models use the same initialization
$\mathbf{x}_0=\tfrac{1}{2}(\mathbf{l}+\mathbf{r})$, which is also IPOPT's, and we evaluate for $30$
steps on the same $500$ held-out test instances using the normalized squared relative error.

\noindent\textbf{Results on Small.} Under identical training and evaluation conditions
on Unstructured Small, SOCPNet with $d{=}256$ (S9) reaches a median $\sre$ of
$3.1\times10^{-6}$ at step $25$, roughly four orders of magnitude below IPM-LSTM
($2.0\times10^{-2}$, Table~\ref{tab:ipmlstm}). Even the much smaller $d{=}64$ model (S8)
outperforms IPM-LSTM by approximately three orders of magnitude ($2.1\times10^{-5}$),
showing that capacity is not the primary driver of the gap.
Fig.~\ref{fig:ipmlstm_overlay} shows that IPM-LSTM's error increases beyond its training
horizon, with a minimum of $1.97\times10^{-2}$ near step $24$. This is precisely the drift
behavior targeted by our stabilization mechanism. These results support our central claim:
the gains come from \emph{what is recursively represented and updated}, a conic graph with a
learned solver.

\noindent\textbf{Results on Medium.} We repeat the same protocol on Unstructured Medium
($n{=}100$), training IPM-LSTM with $T_{\mathrm{outer}}{=}21$ and evaluating $30$ steps on
the same $500$ held-out instances. At the training horizon (step $21$),
Table~\ref{tab:ipmlstm_medium} reports $9.6\times10^{-4}$ for SOCPNet ($d{=}256$) versus
$3.7\times10^{-2}$ for IPM-LSTM. At step $25$, SOCPNet remains at $9.6\times10^{-4}$ versus
$2.5\times10^{-2}$ for IPM-LSTM (about $25\times$). The smaller $d{=}64$ model reaches
$\sre=1.2\times10^{-3}$, still about $20\times$ below IPM-LSTM. Unlike on Small, IPM-LSTM's error
is still decreasing at step $30$, and converges after 50 steps, as shown in Fig.~\ref{fig:ipmlstm_overlay_medium}. SOCPNet remains well below IPM-LSTM throughout.

\noindent\textbf{Complexity.} Per outer iteration, IPM-LSTM is computationally more expensive than SOCPNet ($d{=}256$). On Unstructured Small, the costs are $\sim$100M versus $\sim$40M MACs (multiply-accumulate operations) per outer iteration. On Unstructured Medium, they are $\sim$260M
versus $\sim$150M MACs per outer iteration.

\section{Isolating the Solver-derived Step-size Signal}
\label{app:stepsize}

The leave-one-out ranking of Section~\ref{sec:ablation} drops the step-size component from
an otherwise fixed procedure. Here we retrain the full recipe from scratch without the
solver-supervised step-size target $\mathcal{L}_{\mathrm{step}}$ of Eq.~\eqref{eq:step}, holding
every other loss, feature, and hyperparameter at its S9 value. Both models keep the polar
parametrization of Eq.~\eqref{eq:polar}, so the comparison isolates the solver-derived
information on its own.

\begin{figure}[h!]
  \centering
  \includegraphics[width=0.6\linewidth]{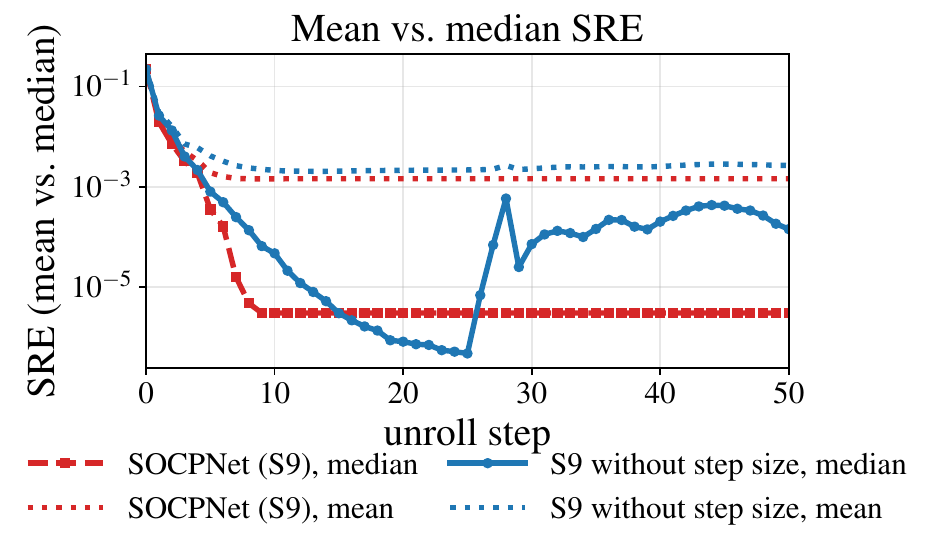}
  \caption{On Unstructured Small: mean and median $\sre$ for S9 with and without step-size control.}
  \label{fig:stepsize-tail}
\end{figure}

\begin{figure}[h!]
  \centering
  \includegraphics[width=0.6\linewidth]{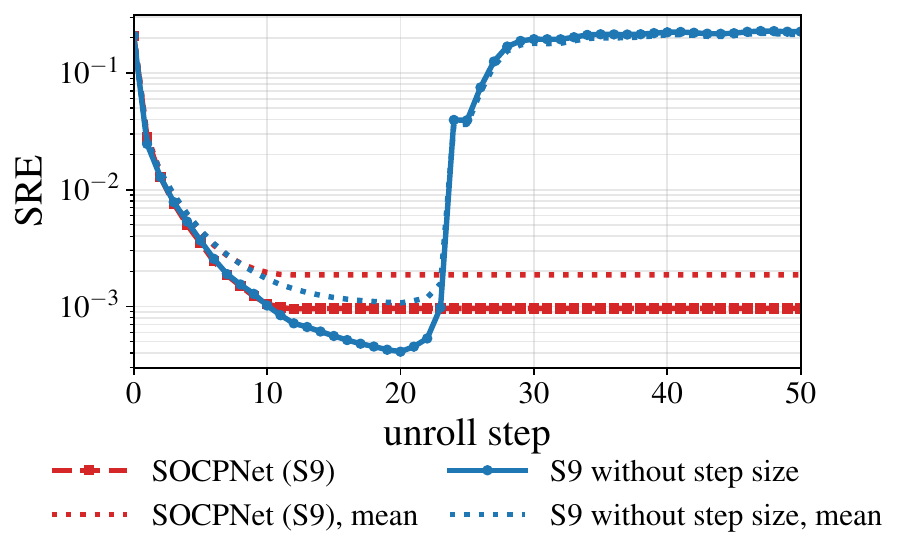}
  \caption{On Unstructured Medium: mean and median $\sre$ for S9 with and without step-size control.}
  \label{fig:stepsize-medium}
\end{figure}

On Unstructured Small, Fig.~\ref{fig:stepsize-tail} shows that the ablated model
attains a median $\sre$ of $4.8\times10^{-7}$, about $6\times$ below S9's
$3.1\times10^{-6}$. Removing the solver signal does not cost median accuracy. Despite the better median, the mean rises to
$2.2\times10^{-3}$ from S9's $1.4\times10^{-3}$, so the mean-to-median ratio grows from
$4.7\times10^{2}$ to $4.5\times10^{3}$. The $90$th percentile grows from $3.2\times10^{-3}$
to $7.1\times10^{-3}$, and the fraction of test instances finishing above $\sre = 10^{-2}$
grows from $4.4\%$ to $7.8\%$. The solver-derived step size therefore acts as a stabilizer
on the worst trajectories. We also observe that without step-size control, the iterates diverge after the training horizon of 25 steps.

On Unstructured Medium, Fig.~\ref{fig:stepsize-medium} shows a similar behavior but with a stronger effect.
Without step-size control, the $\sre$ is further improved below the training horizon, with a median value of $4.1\times10^{-4}$ near step $20$. However, by step $25$ it
has drifted to $3.9\times10^{-2}$, and by step $50$ to $2.3\times10^{-1}$. In contrast, the full S9 recipe stays at $9.6\times10^{-4}$.

\section{Medium-scale Experiments on Other SOCP Families}
\label{app:medium-families}

\begin{figure}[t]
  \centering
  \includegraphics[width=\linewidth]{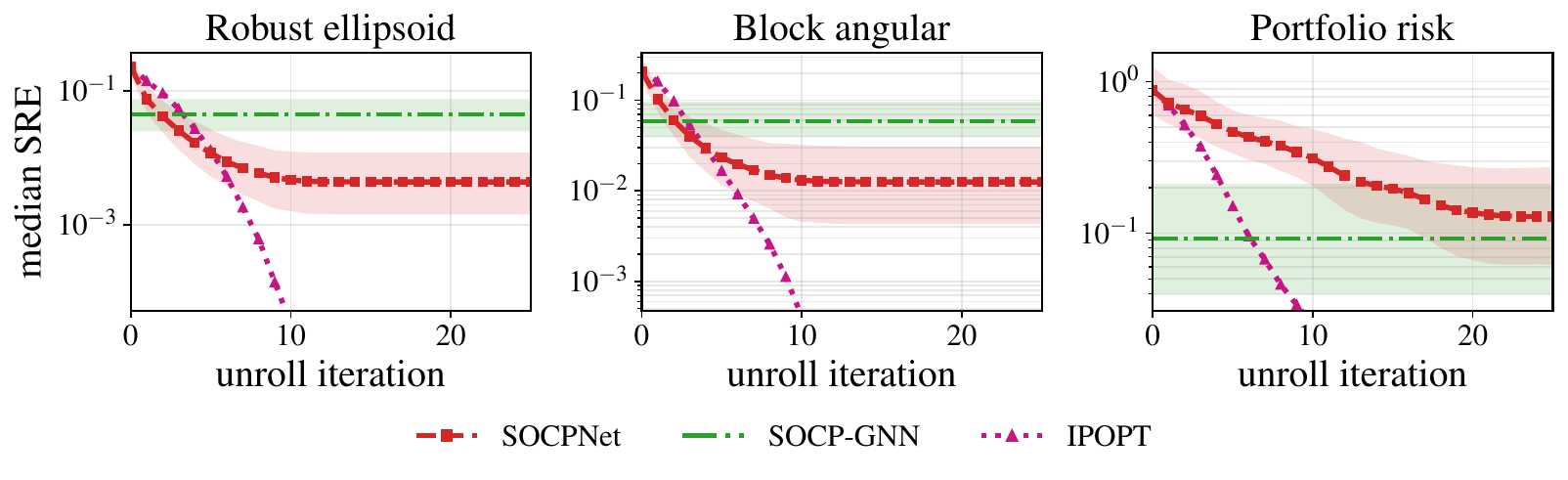}
  \caption{Medium SOCP families: SOCPNet vs.\ SOCP-GNN vs.\ IPOPT.
  As Fig.~\ref{fig:generalization}, at Medium scale ($n{=}100$, $h{=}25$).}
  \label{fig:medium-family-e2e}
\end{figure}

We repeat the per-family comparison of Fig.~\ref{fig:generalization} at Medium scale
in Fig.~\ref{fig:medium-family-e2e} here. In the left and center panels, SOCPNet reaches median $\sre = 4.4\times10^{-3}$ against
$4.4\times10^{-2}$ for the SOCP-GNN on robust ellipsoid ($\sim\!10\times$), and
$1.3\times10^{-2}$ against $5.9\times10^{-2}$ on block angular ($\sim\!4.7\times$). Both
confirm the direction of the Small-scale result, with SOCPNet remaining more accurate than
the SOCP-GNN at Medium scale, though the margin is roughly twice as large on robust
ellipsoid as on block angular, so we do not claim a fixed advantage factor across families.
\emph{Portfolio risk} (right panel) is the exception: SOCPNet plateaus at median $\sre \approx 1.3\times10^{-1}$, slightly above the SOCP-GNN's $9.3\times10^{-2}$. Hence, for this family, the advantage of SOCPNet over SOCP-GNN does not carry over from Small to Medium scale.

\section{Zero-shot Transfer Across Problem Scale}
\label{app:zero-shot}

\begin{figure}[t]
  \centering
  \includegraphics[width=0.8\linewidth]{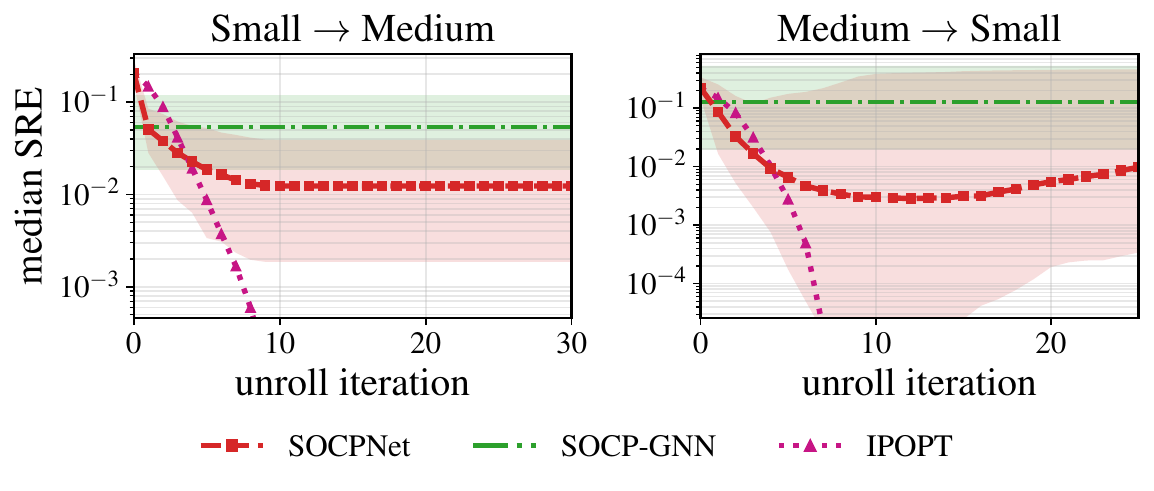}
  \caption{Zero-shot scale transfer, no retraining. Per-step median $\sre$
  ($\downarrow$ better; log axis) for a model trained at one scale and evaluated at the other.
  Same conventions as Fig.~\ref{fig:generalization}.}
  \label{fig:zero-shot}
\end{figure}

Here we keep a \emph{single trained model fixed} and evaluate it on the other problem
scale without retraining or any change to the network. A Small-trained SOCPNet ($n{=}50$) evaluated on Medium instances ($n{=}100$) reaches a
median $\sre$ of $1.24\times10^{-2}$, against $5.34\times10^{-2}$ for the single-shot SOCP-GNN
transferred the same way. In the opposite direction, a Medium-trained SOCPNet evaluated on Small
reaches $9.79\times10^{-3}$ against $1.30\times10^{-1}$, so the recursion keeps a
$4.3-13.2\times$ advantage over the single-shot baseline in both directions
(Fig.~\ref{fig:zero-shot}). Two caveats temper this. Both transferred models remain far from
in-domain accuracy, and the Medium$\to$Small recursion reaches its
best value, $2.8\times10^{-3}$, at step $12$ and then drifts upward, so the transferred update is
not contractive over the full horizon. We therefore report scale transfer as a robustness property
of the shared update. It is no substitute for training at the target size.

\section{Cross-family Generalization: Leave-one-out Testing}
\label{app:loo}

We also evaluate transfer across problem families at Small scale. For each of the four
families, we train on the remaining three and test on the held-out family, under the protocol
of Section~\ref{sec:setup}, and repeat the same folds for the single-shot SOCP-GNN so that both
models see identical data.

\begin{figure}[h!]
  \centering
  \includegraphics[width=0.6\linewidth]{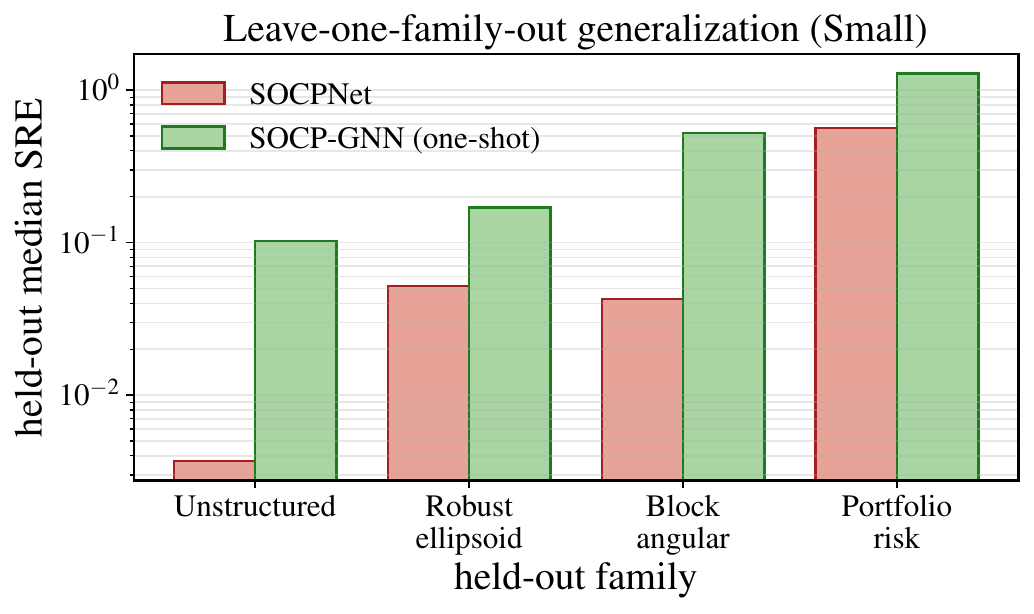}
  \caption{Cross-family leave-one-out at Small scale. Median $\sre$ on the held-out family
  ($\downarrow$ better; log axis) for SOCPNet and the single-shot SOCP-GNN trained on the other
  three families.}
  \label{fig:loo}
\end{figure}

SOCPNet is more accurate than the single-shot baseline on every fold, by $2.3$ to
$27\times$ (Table~\ref{tab:loo}, Fig.~\ref{fig:loo}). Whatever cross-family transfer exists comes
from the recursion, one-shot capacity does not supply it. The absolute numbers are nonetheless
uneven. Held-out Unstructured is the strongest fold at $3.8\times10^{-3}$, comparable to what the
single-shot SOCP-GNN reaches when trained on that family (Table~\ref{tab:main}), though
still three orders of magnitude above in-family SOCPNet. Robust ellipsoid and block angular
fall to around $5\times10^{-2}$, and portfolio risk to $5.7\times10^{-1}$, which is no better than
a coarse approximation. Portfolio
risk is also the family on which SOCPNet fails to improve over the baseline at Medium scale
(Appendix~\ref{app:medium-families}). We therefore do not claim that a single SOCPNet
generalizes across SOCP families, which structural properties transfer is left to future work.

\begin{table}[h!]
  \centering
  \caption{Cross-family leave-one-out at Small scale. Median $\sre$ on the held-out family
  ($500$ test instances) after training on the other three ($\downarrow$ better). The ratio is the single-shot error divided by SOCPNet's $\sre$.}
  \label{tab:loo}
  \small
  \begin{tabular}{lccc}
    \toprule
    Held-out family & SOCPNet & SOCP-GNN (one-shot) & ratio \\
    \midrule
    Unstructured      & $\mathbf{3.77\times10^{-3}}$ & $1.03\times10^{-1}$ & $27\times$ \\
    Robust ellipsoid  & $\mathbf{5.19\times10^{-2}}$ & $1.70\times10^{-1}$ & $3.3\times$ \\
    Block angular     & $\mathbf{4.27\times10^{-2}}$ & $5.25\times10^{-1}$ & $12\times$ \\
    Portfolio risk    & $\mathbf{5.65\times10^{-1}}$ & $1.28\times10^{0\phantom{-}}$ & $2.3\times$ \\
    \bottomrule
  \end{tabular}
\end{table}

\section{Hybrid Warm-Start: Solution Accuracy and Timing}
\label{app:medium-hybrid}

\begin{figure}[h!]
  \centering
  \includegraphics[width=\linewidth]{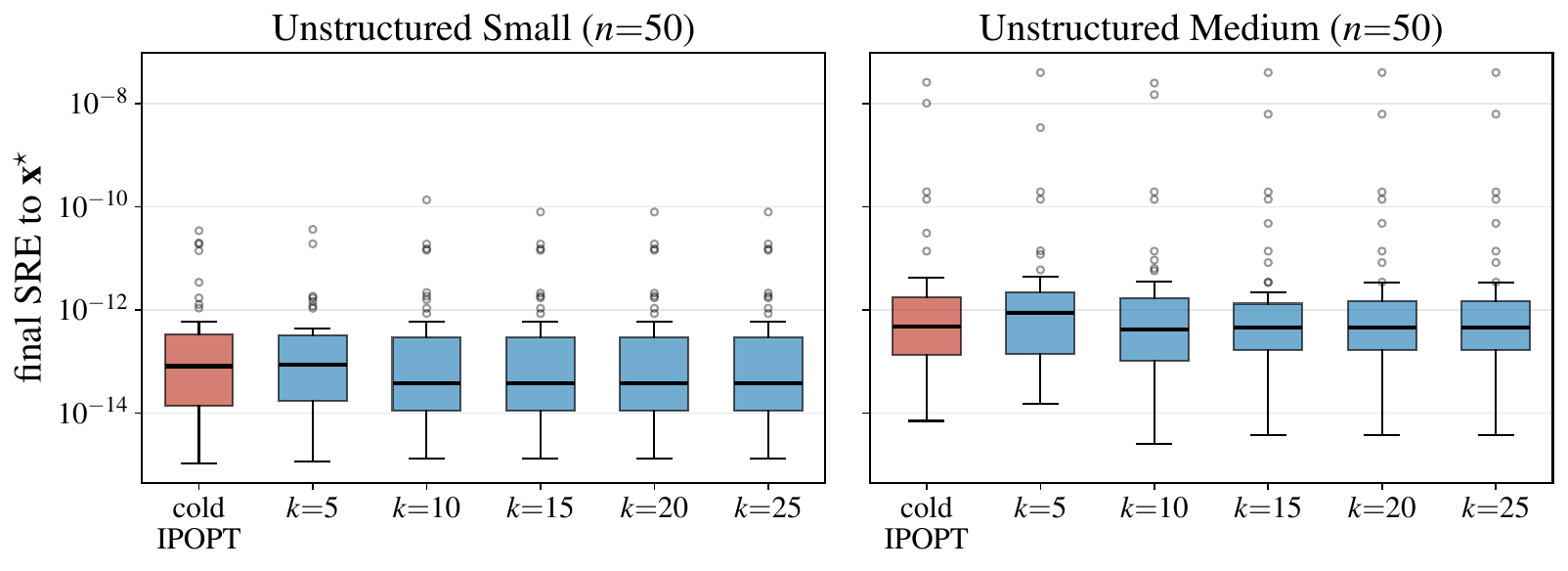}
  \caption{Hybrid solver: final solution accuracy. Final SRE for cold start vs.\ SOCPNet warm start at handoff step $k$, on Unstructured Small (left) and
  Medium (right) ($50$ test problems each, SOCPNet $d{=}256$).}
  \label{fig:hybrid-accuracy}
\end{figure}

Figure~\ref{fig:hybrid-accuracy} reports the final solution accuracy of the hybrid solver
that accompanies the iteration counts of Fig.~\ref{fig:hybrid}. Warm starting does not
compromise accuracy. Indeed, on both the Unstructured Small and Medium datasets, every run (cold
or warm-started at any handoff step $k$) converges to the optimal within the solver's machine-precision
tolerance, $\sre\sim10^{-13}$.

Table~\ref{tab:hybrid-time} gives a detailed breakdown of the wall-clock times. The speed-up is bold where the warm start reduces total time. In Small, a $\sim 2 \times$ speed-up is observed for all $k\geq 5$.
At Medium scale, the iteration saving ($15\to10$ median for $k{\geq}15$) does not result in a significant runtime speed-up because a fixed per-call solver overhead dominates the short forward pass. 

\begin{table}[t]
  \centering
  \caption{Hybrid warm start: iteration and wall-clock breakdown. Medians over
  $50$ test problems per dataset. $k$ is the handoff step, $k{=}0$ the cold-start baseline.
  SOCPNet time is the time spent in the $k$ forward passes, IPOPT the subsequent exact solve,
  and total their sum.}
  \label{tab:hybrid-time}
  \small
  \begin{tabular}{llcccccc}
    \toprule
    & $k$ & $0$ & $5$ & $10$ & $15$ & $20$ & $25$ \\
    \midrule
    \multirow{5}{*}{\rotatebox{90}{Small}}
      & IPOPT iterations   & $11$    & $7$    & $6$    & $6$    & $6$    & $6$ \\
      & SOCPNet time (s)   & $0.00$  & $0.11$ & $0.21$ & $0.32$ & $0.43$ & $0.54$ \\
      & IPOPT time (s)     & $17.00$ & $8.89$ & $7.85$ & $7.71$ & $7.74$ & $7.89$ \\
      & total time (s)     & $17.00$ & $9.00$ & $8.16$ & $8.02$ & $8.17$ & $8.47$ \\
      & speed-up           & $1.0\times$
                           & $\mathbf{1.9\times}$
                           & $\mathbf{2.1\times}$
                           & $\mathbf{2.1\times}$
                           & $\mathbf{2.1\times}$
                           & $\mathbf{2.0\times}$ \\
    \midrule
    \multirow{5}{*}{\rotatebox{90}{Medium}}
      & IPOPT iterations   & $15$   & $12$   & $10.5$ & $10$   & $10$   & $10$ \\
      & SOCPNet time (s)   & $0.00$ & $0.03$ & $0.06$ & $0.09$ & $0.12$ & $0.15$ \\
      & IPOPT time (s)     & $2.18$ & $2.02$ & $2.07$ & $2.06$ & $1.99$ & $2.09$ \\
      & total time (s)     & $2.18$ & $2.05$ & $2.13$ & $2.15$ & $2.14$ & $2.25$ \\
      & speed-up           & $1.0\times$
                           & $1.1\times$
                           & $1.0\times$
                           & $1.0\times$
                           & $1.0\times$
                           & $1.0\times$ \\
    \bottomrule
  \end{tabular}
\end{table}

\section{Reproducibility}
\label{app:runtable}

A code repository accompanying this paper provides all resources required to reproduce the experiments, including the instance generators for all four dataset families, the IPOPT pipeline used to generate the ground-truth trajectories, the SOCPNet implementation and training code, the evaluation and plotting scripts, and the configuration files for all reported runs.
To support reproducibility, we release the source code, data-generation scripts, and experimental setup at \url{https://github.com/nokia/SOCPNet}.

The generation procedure is specified in Appendix~\ref{app:families} for the structured
families and in Section~\ref{sec:setup} together with Appendix~\ref{app:training} for the
Unstructured benchmarks and the IPOPT ground truth. Each figure and table derives from a
single training run whose identifier and configuration are retained in the repository, so
that every reported number can be traced to the checkpoint that produced it. The
configurations record the dataset family and scale, the architecture and ablation variant,
the loss composition, and the optimization settings (batch size, epochs, optimizer,
learning-rate schedule, message passing rounds, hidden dimension, trajectory length, and
random seed). All runs use the fixed seed reported in Appendix~\ref{app:training}.

\end{document}